%% file: main.tex
\documentclass[sigplan, 10pt, twocolumn]{acmart}
\renewcommand\footnotetextcopyrightpermission[1]{}
\usepackage{multirow}
\usepackage{graphicx}
\usepackage{latexsym}
\usepackage{caption}
\usepackage[ruled,linesnumbered]{algorithm2e}
\usepackage{color}
\usepackage[noend]{algpseudocode}
\usepackage{url,epsfig,array}
\usepackage{leftidx}
\usepackage{amsmath}
\usepackage{epstopdf}
\usepackage{cases}
\usepackage{bm}
\usepackage{wasysym}

\usepackage{bbding}

\usepackage{dsfont}
\usepackage{ifpdf}
\usepackage{epsfig}
\usepackage{amsfonts}

\usepackage{amssymb}
\usepackage{paralist}
\usepackage{comment}
\usepackage{xspace}
\usepackage{setspace}
\usepackage{color}
\usepackage{balance}
\usepackage{subfigure}
\usepackage{threeparttable}

\usepackage{enumitem}
\usepackage{makecell}
\usepackage{newtxmath}
\usepackage{tikz}
\newcommand*\circled[1]{\tikz[baseline=(char.base)]{
            \node[shape=circle,draw,inner sep=1pt] (char) {#1};}}

\def\ie{\textit{i.e.}\xspace}

\def\etc{\textit{etc.}\xspace}

\def\eg{\textit{e.g.}\xspace}

\AtBeginDocument{%
  }

\setcopyright{acmcopyright}
\copyrightyear{2025}
\acmYear{2025}
\acmDOI{xxxxxx}
\acmConference[25]{}
\acmISBN{xxxxxx/xx/xx}

\begin{document}
\title{A Robust Federated Learning Framework for Undependable Devices at Scale}


\author{Shilong Wang\textsuperscript{1},~Jianchun Liu\textsuperscript{1},~Hongli Xu\textsuperscript{1}}
\author{Chunming Qiao\textsuperscript{2},~Huarong Deng\textsuperscript{3},~Qiuye Zheng\textsuperscript{3},~Jiantao Gong\textsuperscript{3}}
\affiliation{%
    \institution{\textsuperscript{1}University of Science and Technology of China, China}
    \institution{\textsuperscript{2}State University of New York at Buffalo, USA}
    \institution{\textsuperscript{3}OPPO, China}
  \country{}
}

\begin{abstract}
\input{content/abstract.tex}

\end{abstract}

\ccsdesc[500]{Computing methodologies~Distributed artificial intelligence}
\ccsdesc[500]{Networks~Mobile networks}


\keywords{Federated Learning, Machine Learning Systems, Device Undependability, AIoT}



\maketitle

\section{Introduction}\label{sec_introduction}
\input{content/intro.tex}

\section{Background and Motivation}\label{sec:prelim}
\input{content/background_motivation.tex}
\section{Overview of FLUDE}\label{sec:challenges}
\input{content/system_overview.tex}

\section{Detail Design of FLUDE}\label{sec:design}
\input{content/design.tex}

\section{Performance Evaluation}\label{sec:evaluation}
\input{content/simulation.tex}
\section{Related Work}\label{sec:related}

\input{content/works.tex}

\section{Conclusion}\label{sec:conclusion} 
\input{content/conclusion.tex}



\bibliographystyle{unsrt}
\bibliography{refs}

\appendix

\end{document}

%% file: content/abstract.tex
In a federated learning (FL) system, many devices, such as smartphones, are often undependable (\eg, frequently disconnected from WiFi) during training.
Existing FL frameworks always assume a dependable environment and exclude undependable devices from training,
leading to poor model performance and resource wastage. 
In this paper, we propose FLUDE to effectively deal with undependable environments. 
First, FLUDE assesses the dependability of devices based on the probability distribution of their historical behaviors (\eg, the likelihood of successfully completing training). Based on this assessment, FLUDE adaptively selects devices with high dependability for training. 
To mitigate resource wastage during the training phase, FLUDE maintains a model cache on each device, aiming to preserve the latest training state for later use in case local training on an undependable device is interrupted. 
Moreover, FLUDE proposes a staleness-aware strategy to judiciously distribute the global model to a subset of devices, thus significantly reducing resource wastage while maintaining model performance. 
We have implemented FLUDE on two physical platforms with 120 smartphones and NVIDIA Jetson devices. Extensive experimental results demonstrate that FLUDE can effectively improve model performance and resource efficiency of FL training in undependable environments.

%% file: content/intro.tex
The advent of artificial intelligence (AI), underpinned by deep neural networks \cite{lecun2015deep, khurana2023natural, NEURIPS2023_7058bc19}, 
has significantly enhanced the convenience of people's daily lives. 
The powerful capabilities of AI rely on massive amounts of data for model training.
Traditionally, central servers need to collect a large amount of user data from edge devices to the cloud for training, which significantly increases the risk of data privacy breaches. 
To address the above privacy concerns, federated learning (FL) \cite{huang2024federated, liu2024vertical} is proposed as a new paradigm of model training, which enables edge devices distributed in different geographical locations to collaboratively train a shared model while keeping their data locally. 
The ability to protect data privacy has led to the widespread application of FL in privacy-sensitive scenarios such as e-commerce recommendation \cite{niu2020billion}, smart cities \cite{zheng2022applications}, and healthcare \cite{nguyen2022federated}.\\
\textbf{Challenges of FL Systems.} 
In FL, devices conduct local training and then upload their local models to the central server for global aggregation. Considering a smartphone-based FL system, only those phones (\ie, devices) that satisfy certain specific status conditions, \eg, being WiFi-connected, charging, and having idle CPU/GPU resources, will participate in model training with minimal interference to the user experience \cite{bonawitz2019towards}. We refer to these devices as \textit{dependable} (otherwise \textit{undependable}). In practice, FL often operates in environments with many undependable devices.
For example, devices in FL systems are likely to exit federated training due to unstable network conditions (\eg, WiFi disconnected) or insufficient battery levels. 
In addition, users may run applications (\eg, playing games) on their devices at any time, which will consume considerable CPU/GPU computing resources. In order to avoid model training occupying CPU/GPU resources of running applications, the processes of FL on these devices will be disabled. This undependability of devices is confirmed by a report from Google \cite{bonawitz2019towards}, which indicates that many devices cannot satisfy these conditions of participating in FL.
Furthermore, according to statistics of the real-world FL system of OPPO\footnote{A famous electronics manufacturers and mobile Internet service providers.} for mobile theme recommendation, the average percentage of undependable devices is about 30\%-40\% due to the stringent status conditions required for devices to participate in training ($\S$\ref{subsection:Device-Undependability-in-Federated-Learning}). 
More seriously, these undependable behaviors of devices are highly dynamic and random, and thus cannot be easily predicted.\\
\textbf{Status Quo and Limitations.} While existing FL works \cite{mcmahan2017communication, wu2020safa, lai2021oort, li2021hermes, li2022pyramidfl, sun2022fedsea, wang2022asyncfeded, wang2023bose, cai2023efficient, cai2023federated, zheng2023autofed, xu2024fwdllm} have made significant strides in many aspects, such as reducing communication costs for model transmission, 
they often assume dependable environments by default for FL systems. 
However, FL often encounters device undependability in real-world scenarios. 
One primary concern is that a significant proportion of undependable devices fail to contribute to global model aggregations, which impairs the global model accuracy ($\S$\ref{sec:performance_limitation}). 
This is particularly problematic when some undependable devices hold unique and crucial data. 
Additionally, in undependable situations, the participation frequencies of different devices vary significantly, causing the global model to be biased toward devices with more frequent participation. 
The imbalanced frequencies further compromise the fairness and the global model accuracy ($\S$\ref{sec:performance_limitation}). 
Furthermore, if devices become undependable, their trained intermediate local model will be discarded.
Once these devices participate in training again, they must re-download the global model and restart training from the beginning, resulting in considerable wastage of computational and communication resources ($\S$\ref{sec:resource_limitation}). 
In practice, this will lead to an unaffordable surge in training expenses, especially in a large-scale FL system, rendering FL as a nonviable option.\\
\textbf{Our Solutions.} 
To address the limitations of existing FL frameworks, we propose FLUDE, an efficient FL framework for improving model performance and resource efficiency in undependable environments. 
Specifically, FLUDE first cherry-picks participants for training at the beginning of each training round. However, selecting proper participants is challenging due to the need to consider a confluence of factors, including the varying likelihood of devices becoming undependable and their heterogeneous participation frequencies. Specifically, the participation frequencies of devices vary significantly, with some dependable devices participating in training more frequently than others, leading to a bias in the global model distribution. 
Simply selecting devices with high dependability for training will further exacerbate this imbalance. 
To address this issue, FLUDE first measures the dependability of devices based on the probability distribution of their historical behaviors, and then leverages an online exploration-exploitation strategy to probabilistically select participants with high dependability. Moreover, in order to mitigate bias, FLUDE reduces the priority of a device if its participation frequency exceeds a threshold.
In this way, FLUDE can fully exploit the trade-off between device dependability and participation frequency to improve the training performance of FL systems.

Despite careful participant selection, some devices may still inevitably become undependable during federated training.
To further eliminate the impact of device undependability on federated training, FLUDE configures a local model cache on each device. The cache can preserve the training progress of undependable devices before local training is interrupted.
When those undependable devices become dependable and rejoin the training process, they will either continue from the cached model (if it is not overly stale) or start anew with a fresh global model. 
This approach can significantly mitigate resource wastage in each round. 

However, locally cached models may become stale compared to the latest global model, with varying extents of staleness across devices. 
Training based on overly stale models will lead to a significant drop in model accuracy.
Conversely, distributing the latest global model to all devices, regardless of the freshness of their cached models, would incur substantial bandwidth costs.
Therefore, it is crucial yet challenging to distribute the latest global model to a proper subset of devices, considering the trade-off between communication costs and model accuracy. To address this, FLUDE proposes an adaptive staleness-aware model distribution method, which greedily distributes the latest global model to a subset of devices according to the staleness of their cached local models. Essentially, for undependable devices rejoining the training, FLUDE decouples the training participation from distributing the latest global model.\\
\textbf{Contributions.} 
Overall, we make the following contributions in this paper:
\begin{itemize}[leftmargin=*]
    \item To the best of our knowledge, FLUDE is the first FL framework addressing the key challenges of device undependability in practical systems.
    \item This work qualitatively and quantitatively highlights the impact of device undependability on FL system efficiency and model performance. 
    \item FLUDE contributes several novel techniques, \eg, adaptive device selection, efficient model caching and staleness-aware model distribution, that address the limitations of state-of-the-arts in undependable environments.
    \item We have implemented FLUDE on two physical platforms with 40 OPPO smartphones and 80 NVIDIA Jetson devices, respectively. Compared to state-of-the-arts across various training tasks, FLUDE achieves 2.28\%\textemdash7.43\% higher model accuracy, improves time-to-accuracy by 1.2$\times$\textemdash3.2$\times$, and reduces communication costs by 23.71\%\textemdash40.71\%.
\end{itemize}

In the following, \S \ref{sec:prelim} motivates the design of FLUDE by revealing the damaging effects of device undependability. \S \ref{sec:challenges} presents the overview of FLUDE and \S \ref{sec:design} presents the detailed system design. \S \ref{sec:evaluation} introduces the system implementation and reports the evaluation results. Related works are discussed in \S \ref{sec:related}. Finally, \S \ref{sec:conclusion} concludes the paper with future directions.


  



%% file: content/background_motivation.tex
We commence with key observations about device undependability in a real-world FL system. Subsequently, we delineate the challenges that device undependability poses in FL, motivating the design of our proposed system.
\subsection{Key Observations in FL Systems}\label{subsection:Device-Undependability-in-Federated-Learning}
In traditional FL systems, after local training, each trained model on the device will be uploaded back to the central server for global aggregation \cite{mcmahan2017communication, zhao2018federated}.
However, our empirical observations from the real-world FL system\footnote{The mobile theme recommendation system deployed by OPPO.} reveal that a non-trivial portion of devices fail to upload their local models to the central server in each round.
This is predominantly attributed to the \emph{undependable} nature of these devices. The model training on undependable devices that habitually fail to satisfy the status conditions for model training, including but not limited to, having idle GPU/CPU resources, being WiFi-connected and charging, is highly susceptible to interruption.
Consequently, massive undependable devices may dynamically and frequently exit federated training, obstructing the operation of FL systems.

According to key observations from the real-world FL system, there is a significant proportion of undependable devices within the system. Specifically, 
the proportion of undependable devices, called \emph{undependability rate},
remains consistently high throughout the FL life cycle, reaching about 30\%-40\%. Besides, according to Google, more than half of the devices cannot meet the status conditions for training \cite{bonawitz2019towards}. As these conditions become stricter, more devices will even become undependable. Moreover, our observations indicate that 
the participation frequencies among devices differ greatly, with some dependable devices participating more than ten times as often as others, introducing a bias into the global model aggregation.
The pervasive undependability across substantial devices poses a severe threat to the operational efficacy and resource efficiency of FL. 

\subsection{Limitations of Existing FL Systems}\label{subsection: motivation}
We delve into a comprehensive analysis on how device undependability detrimentally affects FL performance, which highlights the key limitations of existing works.
Specifically, we conduct experiments by training a 5-layer Convolutional Neural Network (CNN) model \cite{krizhevsky2012imagenet} on the CIFAR-10 dataset \cite{krizhevsky2009learning}, which contains 10 classes of data, in both undependable and dependable environments. 
We set six undependability rates (\eg, 10\%, 20\%, $\cdots$, 60\%) to simulate different undependable environments.
Besides, to simulate different participation frequencies among devices, we model the heterogeneous likelihoods of devices becoming undependable using either a normal distribution (\ie, Undepend.+Normal) or a uniform distribution (\ie, Undepend.+Uniform) with a variance of 0.04, aligning with our empirical observations ($\S$\ref{subsection:Device-Undependability-in-Federated-Learning}). 
Conversely, in the dependable environment (\ie, Depend.), all devices can successfully complete training and upload the local model to the server.
In each round, the central server randomly selects 50 devices from a pool of 250 devices for model training. All experiments train the model on the non-Independent and Identically Distributed (non-IID) data, where each device holds 2 classes of data \cite{li2021hermes}.\\
\subsubsection{Limitations in Model Performance}\label{sec:performance_limitation}

\begin{figure}[t]\centering
    \begin{minipage}[t]{0.438\linewidth}\centering
        \subfigure[Test accuracy \emph{vs.} undependability rate.]{\centering
                \label{fig:compare_acc-1}
                \includegraphics[width=0.95\linewidth]{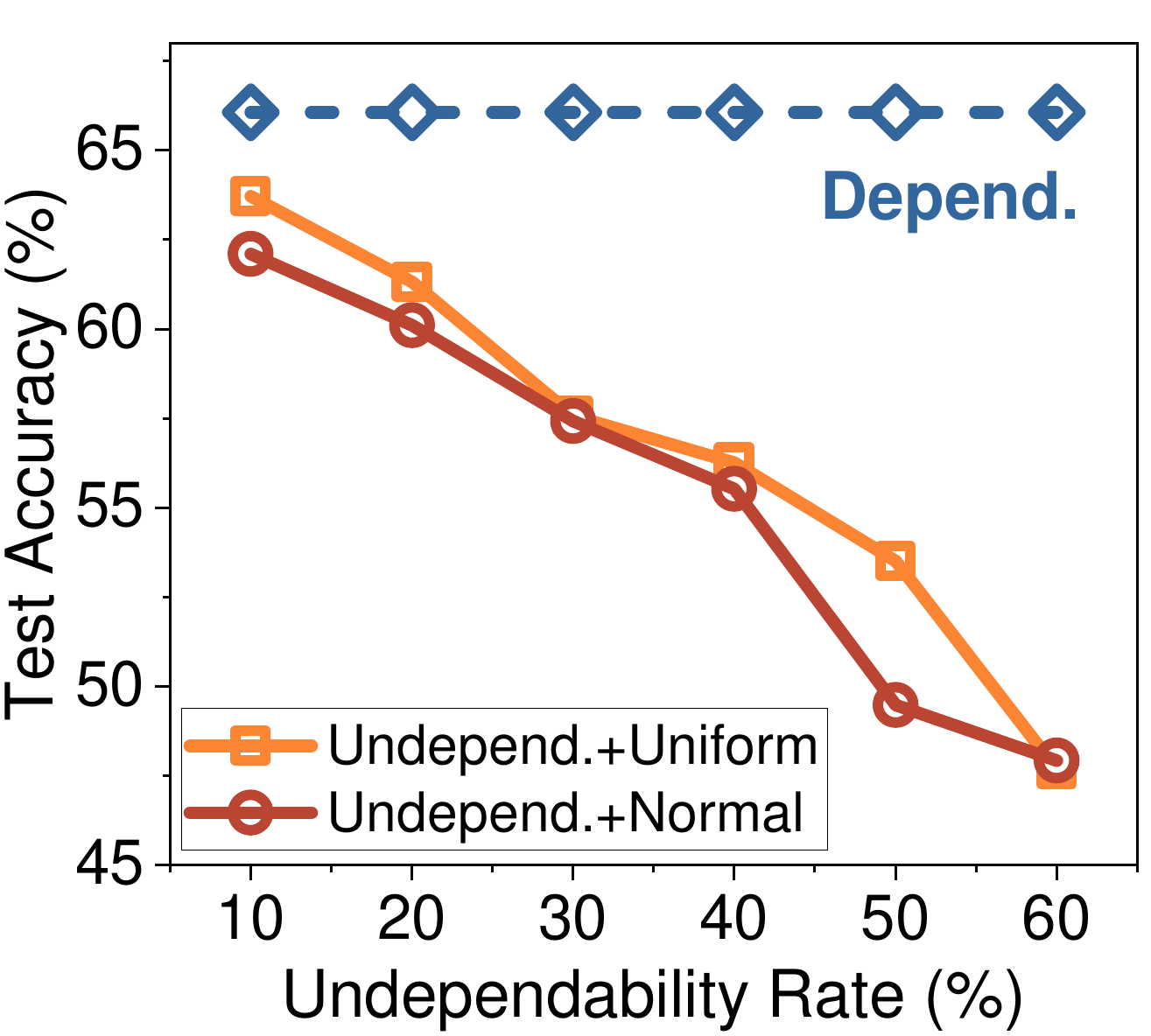}
            }
    \end{minipage}
    \begin{minipage}[t]{0.55\linewidth}\centering
        \subfigure[Test accuracy \emph{vs.} volumes of data across data classes.]{\centering
                \label{fig:compare_acc-2}
                \includegraphics[width=0.955\linewidth]{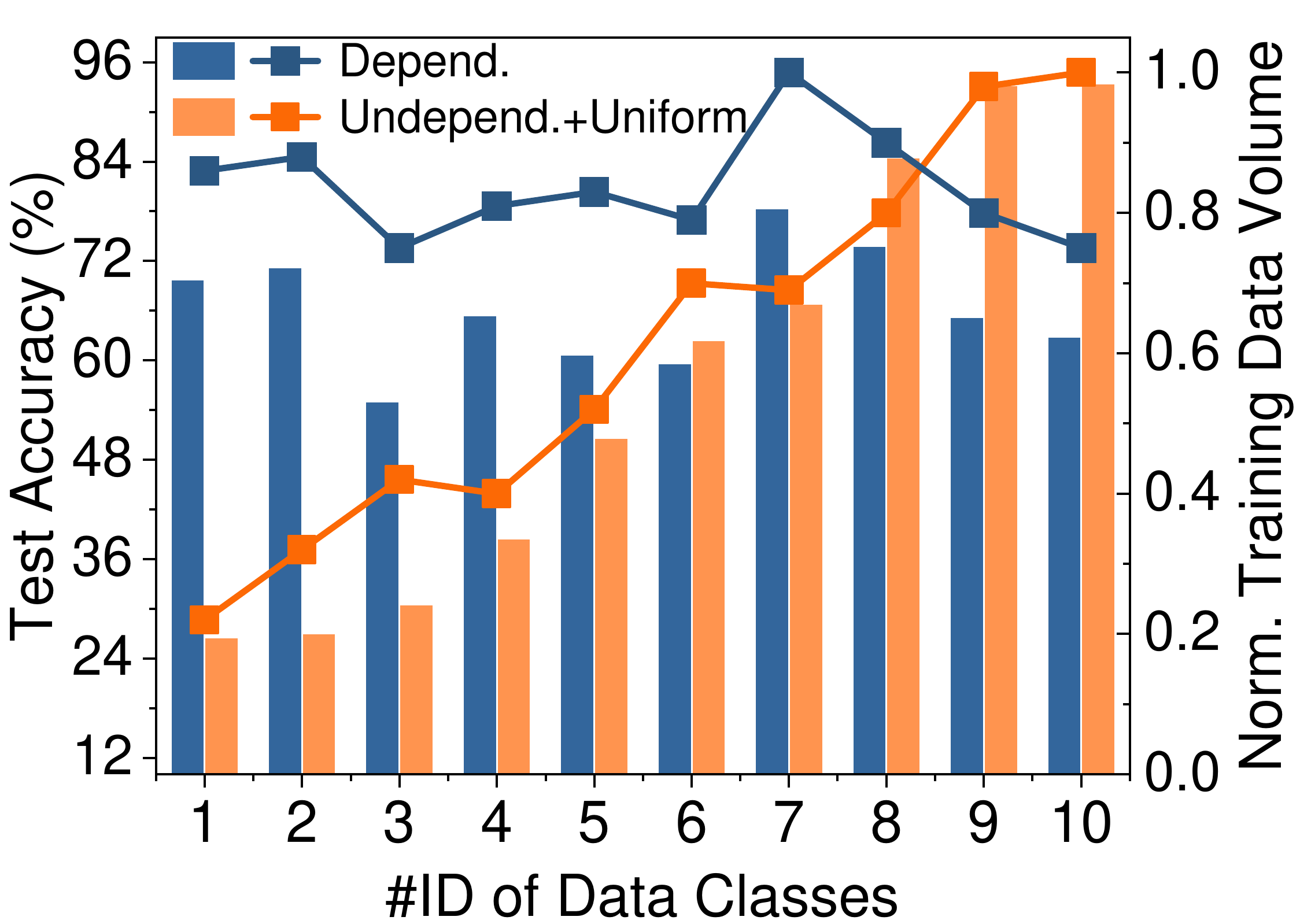}
            }
    \end{minipage}
    \begin{minipage}[t]{1.0\linewidth}\centering
        \subfigure[Test accuracy \emph{vs.} participation frequency across devices. We report
results on a random subset of 50 devices.]{\centering
                \label{fig:compare_acc-3}
                \includegraphics[width=1.0\linewidth]{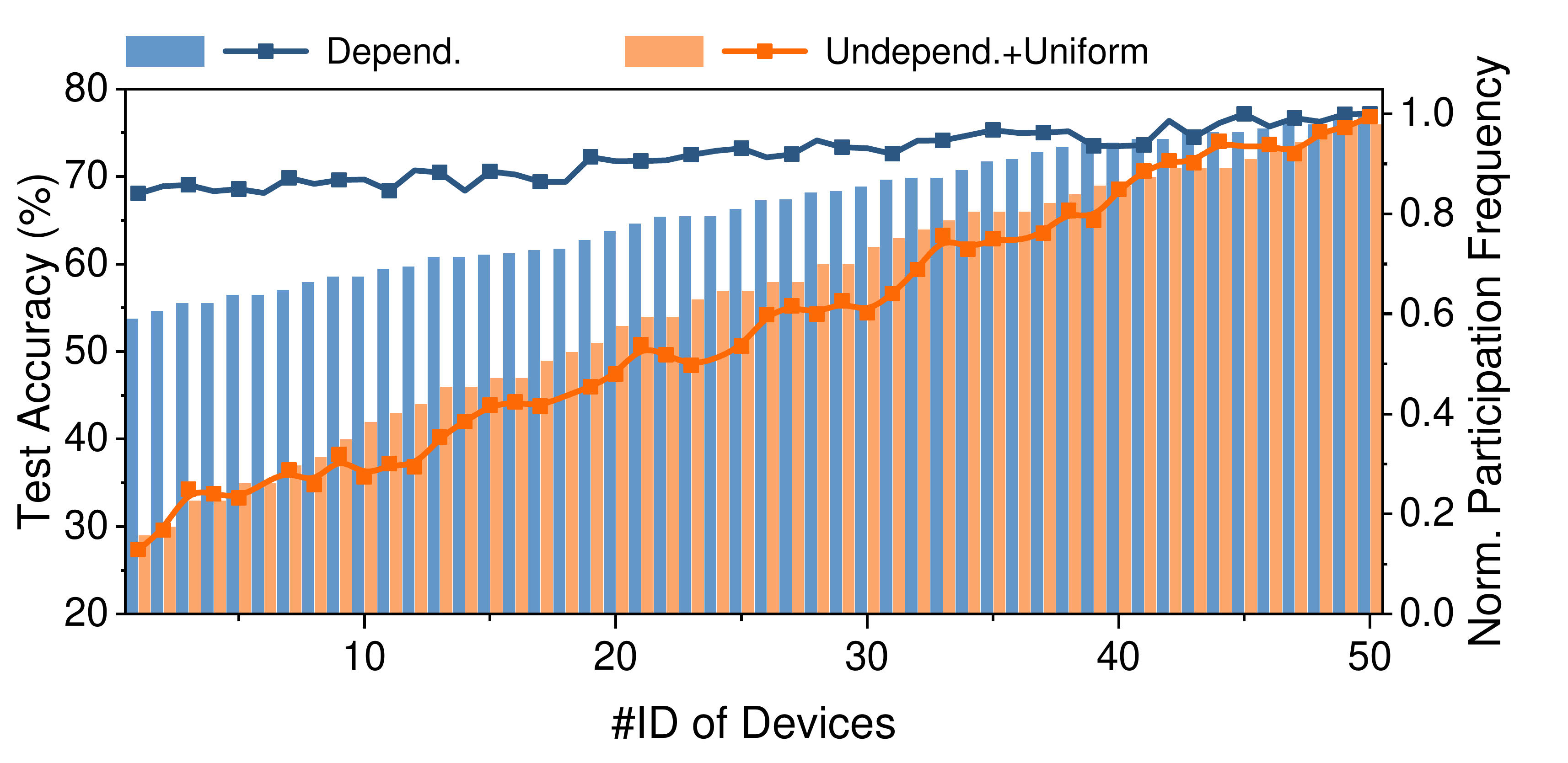}
            }
    \end{minipage}
    \caption{Training performance comparison under dependable and undependable environments. (a) The 
    global model accuracy. (b) The model accuracy (bars) and volumes of data involved in federated training (lines) across data classes. (c) The model accuracy (bars) and participation frequency (lines) across devices.}\label{fig:compare_acc}
\end{figure}
After 500 training rounds, we evaluate the global model's accuracy using the global test dataset, which aggregates the local test dataset from all 250 devices.
As depicted in Figure \ref{fig:compare_acc-1}, the model accuracy suffers a significant degradation in the undependable environments, compared to that in the dependable environment.
For instance, even with a moderate undependability rate of 30\%, the model accuracy in Undepend.+Normal and Undepend.+ Uniform is 8.64\% and 8.49\% lower than that in Depend., respectively. 
The reason is that undependable devices in FL fail to upload their local models to the central server, which leads to a substantial reduction in the volume of local updates available for global aggregation. 
Consequently, the global model's ability to learn from sufficient local data is hindered. 
Besides, the discrepancy in participation frequencies across devices introduces a bias into the global model. This bias manifests in two ways. 
First, it overlooks potential contributions from devices that possibly possess unique and critical data but seldom or never participate in training. 
Second, devices that contribute more frequently to the global model always achieve higher model accuracy, as their local training is more thorough than others. 

Figures \ref{fig:compare_acc-2} and \subref{fig:compare_acc-3} report the global model accuracy across different data classes and devices with an undependability rate of 40\%.
Specifically, after 500 training rounds, we evaluate the accuracy of the global model on each class within the global test dataset, as well as on each device's local test data. Then, we label the data classes (from 1 to 10) and devices (from 1 to 50) according to their ascending accuracy. 
In the undependable environment, the global model performs better (\eg, 93.2\% and 93.4\%) on data classes (\eg, \#\{9, 10\}) with more data involved in federated training. However, due to less participation of data from classes \#\{1, 2, 3\}, the global model suffers from significantly low model accuracies of 26.5\%-30.5\% on these classes. 
Moreover, the devices (\eg, \#[40-50]) with higher participation frequencies can achieve test accuracies of 69.12\%-76.23\%. 
Conversely, devices \#[1-10], which barely participate in training, exhibit relatively poor test accuracy of 29.53\%-42.67\%. 
These results demonstrate that the global model trained by the existing FL system in undependable environments cannot effectively generalize across data from all devices and classes.\\
\subsubsection{Limitations in Resource Efficiency}\label{sec:resource_limitation}
Undependable devices introduce significant resource inefficiency in the following aspects.
(1) When the central server distributes the global model to these undependable devices, it generates substantial unnecessary communication costs, as these devices do not contribute to the global model update. 
(2) Undependable devices must discard their completed training progress and restart local training from scratch upon reselection. 
This cycle wastes considerable computing resources and leads to time inefficiency. 
(3) Device undependability will lead to model performance degradation, as demonstrated in \S \ref{sec:performance_limitation}. 
Therefore, undependable environments require significantly more training rounds and resource costs to reach the target model accuracy compared to dependable environments.
\begin{figure}[t] \centering
    \includegraphics[width=0.41\textwidth]{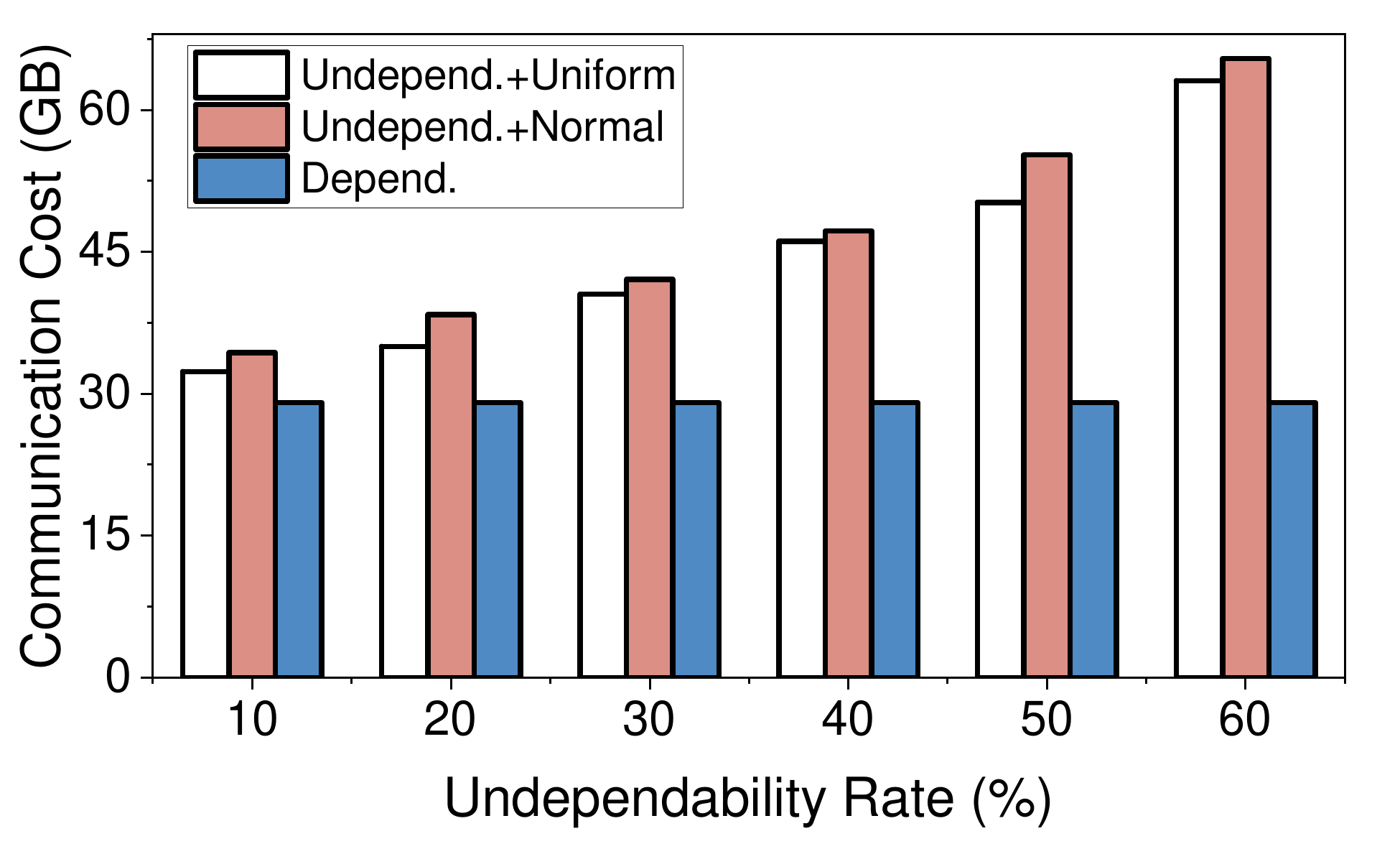}
    \caption{Communication costs to reach the target accuracy of 45\% for training CNN on CIFAR-10.}\label{fig:commun}
\end{figure}

In the experiment, we measure the total communication costs required for federated training to achieve a target accuracy of 45\% (\ie, the minimum achievable accuracy in all environments) under different settings. 
As shown in Figure \ref{fig:commun}, the communication costs exacerbate with the increasing undependability rate. 
For example, given the undependability rate of 30\%, the communication cost of training with Undepend.+ Uniform and Undepend.+Normal increase by 41\% and 45\%, respectively, compared to training with Depend.. 
In extremely undependable environments with an undependability rate of 60\%, communication costs increase significantly by 117\% and 121\%, respectively. Consequently, the increased resource costs will undermine the overall efficiency of FL systems.

In summary, traditional FL frameworks often overlook the device undependability prevalent in real-world scenarios. This oversight results in degraded training performance and resource inefficiency in undependable environments, underscoring the urgent need for a robust FL framework.

%% file: content/system_overview.tex
Motivated by the aforementioned limitations, we propose FLUDE, a novel and efficient FL framework that simultaneously (1) migrates the model performance degradation; and (2) improves the resource efficiency, in realistically undependable environments.
To achieve the above goals, FLUDE judiciously selects participants at the beginning of each training round (\S \ref{subsec:selection}). Specifically, FLUDE prioritizes devices with high dependability according to their historical behaviors to participate in training in each round. This ensures that more dependable devices contribute to the training process, which in turn facilitates more local updates and less resource wastage for global aggregation in each round. 

To further mitigate the damage of device undependability to model training, FLUDE introduces a local model caching mechanism (\S \ref{subsection: Caching}), which preserves the latest training state (\eg, model parameters and learning rate) on undependable devices before training interruption.
This strategy prevents the need to restart training from the latest global model of undependable devices, thereby significantly increasing the likelihood of successful training without unnecessary resource wastage. Based on model caching, the central server carefully distributes the latest global model to \emph{a subset of the selected devices} (\S \ref{subsec:distribution}). The reason has two folds. First, distributing the latest global model to all devices will introduce considerable communication costs, especially in large-scale FL systems. On the contrary, all devices conducting local training based on the locally cached models, which may be stale compared to the latest global model, will introduce errors into the global model and impair the model accuracy. Therefore, FLUDE adaptively distributes the latest global model to only a subset of devices according to the extent of local model staleness, achieving a sweet spot in the trade-off between model performance and communication costs.
\begin{figure}[t]
    \centering
    \includegraphics[width=0.46\textwidth]{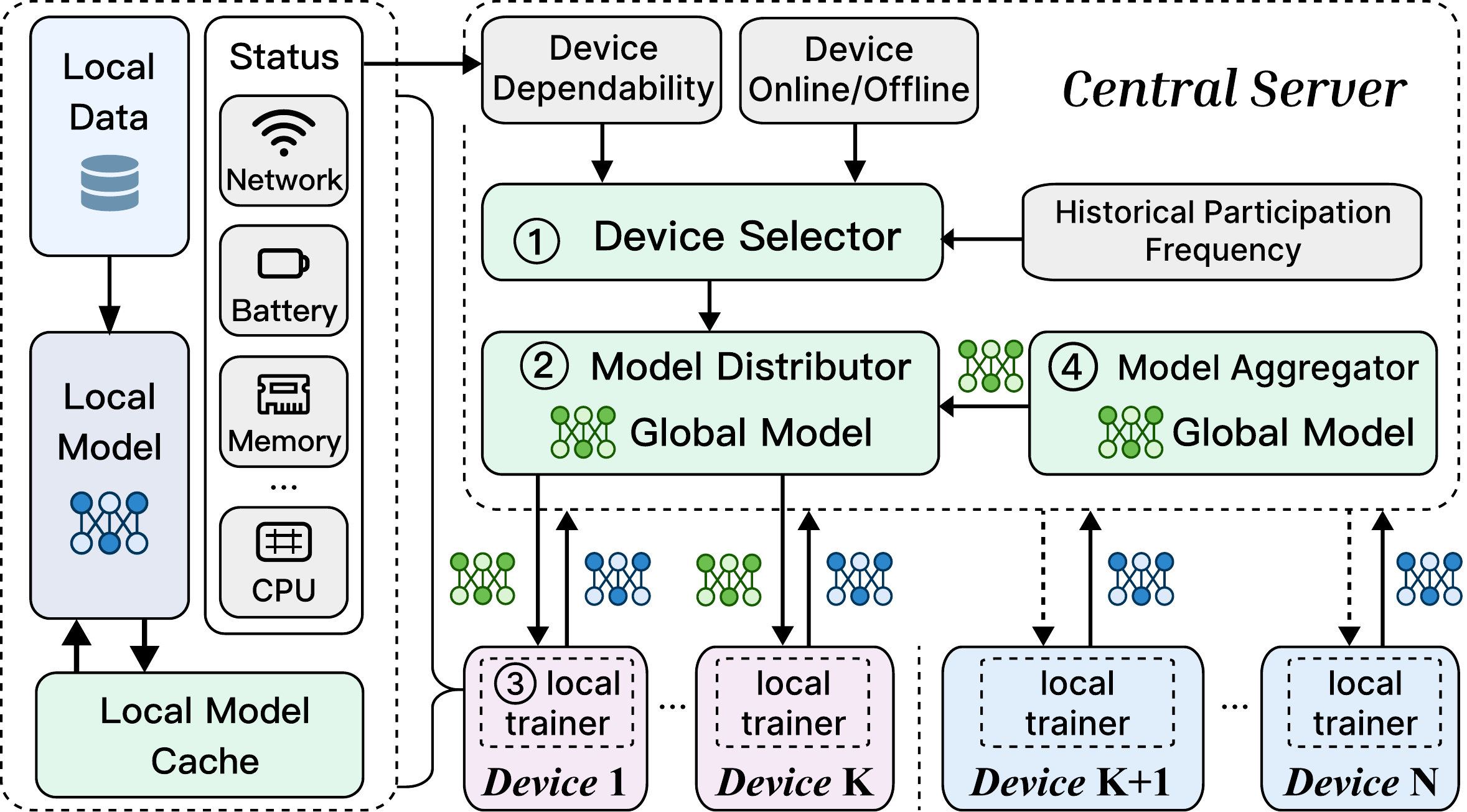}
    \caption{Overview and workflow of FLUDE.}
    \label{fig:overview-of-framework}
\end{figure}

Figure \ref{fig:overview-of-framework} illustrates the overall design and workflow of FLUDE.
At the beginning of each round, the central server employs a device selector (\circled{1}) to identify a set of dependable devices to participate in training, according to the dependability, online status and historical participation frequency of devices.
Next, the model distributor (\circled{2}) in FLUDE decides whether the latest global model is distributed to each selected device or not, considering the trade-off between model accuracy and communication costs.
Upon initiating the training round, local trainers (\circled{3}) of selected devices commence local training , which is based on either the latest global model or their locally cached models. These devices periodically update their local model caches and report their status during local training to the central server. Once local training is completed, the local model parameters are uploaded to the central server. At the end of each training round, the model aggregator (\circled{4}) conducts model aggregation on the received local models to update the global model.


%% file: content/design.tex
\subsection{Adaptive Device Selection}\label{subsec:selection}
A large-scale FL system typically involves numerous devices. 
However, only a small fraction of devices can participate in each training round due to the significant overhead of aggregating local updates from such a huge fleet \cite{bonawitz2019towards}. 
Given the undependable nature of devices, 
selecting which ones to participate in training is crucial for both global model performance and system efficiency. 
A natural solution is to select highly dependable devices for each training round. However, this solution encounters three challenges in practice: 
\begin{itemize}[leftmargin=*]
    \item In real-world FL systems, the undependability of devices can fluctuate over time. \emph{How can the server effectively and efficiently assess the devices' dependability?}
    \item Simply selecting devices with high dependability will lead to more unbalanced participation frequencies among devices than random selection. \emph{How to balance the frequencies between devices with varying levels of dependability?}
    \item A device's dependability can only be evaluated after it has participated in at least one training round. \emph{How to select devices without having to try all devices once?} 
\end{itemize}
To tackle these challenges, we develop an adaptive strategy for participant selection, as described in Algorithm \ref{alg:selection}.\\
\textbf{Device dependability assessment.} To measure the device dependability without incurring excessive overheads, FLUDE assesses a device's dependability based on its historical behavior (Line \ref{line:dependability}).  
Specifically, we define the dependability $R(i)$ of device $i$ as the likelihood of it successfully completing the federated training.
The basic assumption is that $\{R(i), \forall i\}$ follow a specific probability distribution (\eg, normal distribution or beta distribution), which helps predict the dependability of each device. 
An appropriate distribution for $\{R(i), \forall i\}$ can be efficiently chosen by performing statistical analysis on a small random subset of devices (\eg, 100 out of 1,000) over several training rounds (\eg, 5), without incurring significant costs. 
In this context, we use the beta distribution $Beta(\alpha, \beta)$ \cite{nadarajah2007multitude}, characterized by two parameters $\alpha$ and $\beta$, as an example to illustrate how FLUDE dynamically predicts the dependability of each device. 
\begin{algorithm}[t]
	\caption{Adaptive device selection in FLUDE.} \label{alg:selection}
    
    \SetAlgoNoEnd
    \SetAlgoNoLine
    
	\KwIn{Online devices $\mathbb{D}$, explored devices $\mathbb{C}$, penalty $\sigma$, participants size $X$, exploration factor $\epsilon$}
 
	\KwOut{Selected participants $\mathbb{S}$}
 
    \SetKwFunction{Select}{SelectParticipant}
    \SetKwProg{Fn}{Function}{:}{}
    \Fn{\Select{$\mathbb{D}$, $\mathbb{C}$, $X$, $\epsilon$, $\sigma$}}{ 
        $\mathbb{P}\leftarrow\emptyset${\color[RGB]{148,0,211}\footnotesize{\tcp*{Device priorities}}} 
        {\color[RGB]{148,0,211}\footnotesize{\tcc{Update frequency threshold by Eq. \eqref{eq:frequency}}}}
        $Q\leftarrow$ UpdateFreqThreshold()\;\label{line:frequency}
        {\color[RGB]{148,0,211}\footnotesize{\tcc{Calculate device priorities.}}}
        \For{device $i\in\mathbb{C} \cap \mathbb{D}$}{
            {\color[RGB]{148,0,211}\footnotesize{\tcc{Update dependability by Eq. \eqref{eq:dependability}.}}}
            $R(i)\leftarrow$ UpdateDeviceDependability($i$)\;\label{line:dependability}
            $P(i)\leftarrow R(i) \times (\frac{Q}{q_{i}})^{\vmathbb{1}(Q<q_{i})\times\sigma}$\;\label{line:frequencys}
            $\mathbb{P}\leftarrow\mathbb{P}\cup\{P(i)\}$\;\label{line:frequencye}
        }
        {\color[RGB]{148,0,211}\footnotesize{\tcc{Exploit $(1-\epsilon)\times X$ devices.}}}
        $\mathbb{P}\leftarrow$ SortPriority($\mathbb{P}$)\;\label{line:exploitation-explorations}
        $\mathbb{S}\leftarrow$ ExploitDevice($\mathbb{C}$, $\mathbb{P}$, $(1-\epsilon)\times X$)\; \label{line:exploitation}
        {\color[RGB]{148,0,211}\footnotesize{\tcc{Explore $\epsilon\times X$ devices.}}} 
        $\mathbb{O}\leftarrow$ ExploreDevice($\mathbb{D}-\mathbb{C}$, $\epsilon\times X$)\; \label{line:explore}
        $\mathbb{S}\leftarrow\mathbb{S}\cup\mathbb{O}$; 
        $\mathbb{C}\leftarrow\mathbb{C}\cup\mathbb{O}$\;\label{line:exploitation-exploratione}
        \textbf{Return} $\mathbb{S}$\;
    }
\end{algorithm}

Specifically, we initially designate the prior knowledge of an arbitrary device participating in training for the first time as a relatively neutral Beta distribution, such as $Beta(\alpha= 2,\beta= 2)$ in FLUDE. This distribution represents an unbiased starting point, assuming that the device is neither particularly dependable nor undependable. Additionally, the initial values of $\alpha$ and $\beta$ for new devices can be determined based on the average performance of similar devices, manufacturer-provided statistics, or expert judgment.
Through Bayesian theorem \cite{bernardo2009bayesian}, the dependability of device $i$ is updated after observing a new training success or failure of device $i$ by updating $\alpha$ and $\beta$ as follows:
\begin{equation}\label{eq:dependability}
    \begin{cases}
    \alpha_{new} = \alpha + s\\
    \beta_{new} = \beta + f\\
    \mathbb{E}[R(i)] = \frac{\alpha_{new}}{\alpha_{new}+\beta_{new}}
    \end{cases}
\end{equation}
where $s$ is the number of successful training completions by device $i$, and $f$ is the number of failures.
This iterative update results in a refined Beta distribution, \ie, $Beta(\alpha_{new}, \beta_{new})$, whose expected value provides an updated estimate of device $i$'s dependability.\\
\textbf{Balancing participation frequency of devices.} Merely selecting devices based on high dependability would exacerbate disparities in participation frequency across devices. For example, devices with the highest dependability might be selected in every training round, whereas those assessed as less dependable could be consistently overlooked. Consequently, cherry-picking participants for better model performance requires a balanced consideration of both dependability and participation frequency. As such, we formulate the selection priority of device $i$ by associating its dependability with a penalty that accounts for excessive participation frequency:
\begin{equation}\label{eq:priority}
    P(i) = R(i) \times (\frac{Q}{q_{i}})^{\vmathbb{1}(Q<q_{i})\times\sigma}
\end{equation}
where $q_{i}$ is the number of times device $i$ has participated in training, $Q$ is a threshold for participation frequency, and $\vmathbb{1}(x)$ is an indicator function that takes value 1 if $x$ is true and 0 otherwise. In this way, the priority of those devices with excessively high participation frequencies are penalized by a factor $\sigma$. To ensure a more uniform participation frequency among all devices, we define the threshold for participation frequency as the average frequency when devices are randomly selected in each round:
\begin{equation}\label{eq:frequency}
    Q = \frac{\sum_{k=1}^{K}|\mathbb{S}_{k}|}{|\mathbb{A}|}
\end{equation}
where $K$ is the current round number, $\mathbb{S}_{k}$ are selected participants in round $k$ and $\mathbb{A}$ is the set of all devices in FLUDE.\\
\textbf{Exploitation-exploration of dependable devices.} In order to effectively select devices before fully assessing the dependability of all devices, 
FLUDE models the device selection process as a multi-armed bandit problem \cite{auer2002finite}, with the primary objective of balancing the exploration of new devices against the exploitation of known dependable participants (Lines \ref{line:exploitation-explorations}-\ref{line:exploitation-exploratione}).
Specifically, in each round, the server carefully exploits $1-\epsilon$ fraction of devices with top priority to participate in training (Lines \ref{line:exploitation-explorations}-\ref{line:exploitation}), while randomly exploring $\epsilon$ fraction of previously unselected devices (Line \ref{line:explore}). 
Additionally, one can also explore new devices characterized by low CPU/GPU usage, high battery level, \etc, hypothesizing that these devices are more dependable.

\subsection{Local Model Caching}\label{subsection: Caching}
Despite careful participant selection, some devices may still inevitably become undependable during the training phase.
Traditional FL frameworks often discard the trained local models from undependable devices, resulting in considerable losses in completed training progress and elevating communication costs due to the need for redistributing models. 
This practice not only amplifies resource consumption but also reduces the number of local updates received by the server, thereby degrading overall training performance.

To address the challenges posed by device undependability, FLUDE implements a local model caching mechanism, where each device maintains a model cache to periodically save the current training state during local training. 
This mechanism ensures that the completed training progresses on undependable devices are preserved, instead of being discarded directly.
When an undependable device which is halted during the previous local training, becomes dependable and hence participates in training again, it can choose to continue training from the last cached training state, rather than restarting with the latest global model. 
For instance, consider a device with $N$ training samples, of which 0.7$N$ had been processed by local training before this device became undependable. 
In the subsequent round, if the device participates in training again, it can perform local training based on the cached checkpoint instead of downloading the fresh global model, requiring only 0.3$N$ samples to be processed. This model caching significantly reduces communication costs associated with redistributing the global model to undependable devices. 
Furthermore, by allowing devices to build upon their completed training progress, the training efficiency (\eg, shorter computing time and less power consumption) is enhanced and the likelihood of successful training on devices is increased. 
Consequently, the server benefits from receiving a greater volume of local updates in shorter rounds, thereby improving convergence speed.\\
\textbf{Adjusting caching frequency.} In FLUDE, each device periodically caches its training state at a certain frequency. The basic caching frequency (\eg, every minute) is typically set by the central server. 
However, the final caching frequency on each device is adaptive, considering various device-specific factors such as real-time battery level and network stability.
For example, with a lower battery level and less network stability, the device will adjust its caching frequency (\eg, every 30 seconds) to be higher than the basic frequency, accounting for a higher likelihood of becoming undependable. 
Conversely, in a more dependable environment, the device can reduce the caching frequency (\eg, every 5 minutes) to save the overhead required for caching.\\
\textbf{Minimizing caching costs.} To reduce caching costs and ensure efficient storage management, FLUDE employs a rolling mechanism for model caching. 
Specifically, only the latest training state is retained in the cache, while the older one is discarded upon the arrival of a new one.
Consequently, the caching costs of typical models (\eg, CNNs \cite{krizhevsky2012imagenet}) are negligible (typically less than 50MB) compared to the ample storage space (\eg, 64-256GB) of modern mobile devices.



\subsection{Staleness-Aware Model Distribution}\label{subsec:distribution}
Upon participant selection, the central server judiciously distributes the latest global model to a subset of selected devices for local training. The selected participants in each round can be divided into two groups: the first, denoted as $\mathbb{U}$, encompasses devices that either successfully completed local training in their last participation round or have never been selected before. The second, denoted as $\mathbb{V}$, comprises undependable devices that fail to upload their local models to the central server in their last involvement. It is incumbent upon the server to distribute the latest global model to each device within $\mathbb{U}$. Conversely, devices in $\mathbb{V}$ have retained locally cached models, so they can opt to resume training from their cached training state to conserve communication and computational resources. 
However, locally cached models are stale compared to the latest global model. We define the local model's staleness on device $i$ as the discrepancy between the round when caching the model and the current round. 
Training with these stale models will introduce errors into the global model \cite{zhou2022towards, damaskinos2022fleet, liu2021fedpa}. 
On one hand, updating all devices in $\mathbb{V}$ with the newest global model could incur substantial communication costs. 
On the other hand, allowing some devices in $\mathbb{V}$ to train with significantly stale models could lead to notable declines in model accuracy \cite{miao2023robust, wu2023hiflash}. Therefore, it is crucial for the server to determine which devices in $\mathbb{V}$ should conduct local training based on the latest global model versus continuing from their cached model.

A straightforward and effective method for model distribution is to distribute the global model to devices with overly stale local models. Specifically, devices in $\mathbb{V}$ with cached local models exceeding a staleness threshold $W$ are required to receive the latest global model from the central server. 
However, determining a proper value for the threshold poses a challenge: an excessively large $W$ compromises model accuracy, while an overly small $W$ incurs unnecessary communication costs. To seek a sweet spot in the trade-off between model performance and communication costs, FLUDE develops an adaptive strategy for model distribution.
In each round, after selecting participants, each selected device reports its caching status (\eg, whether it has a cached local model) to the server. Then the server calculates the staleness of cached local models for each device in $\mathbb{V}$ as well as their average staleness $H$. According to $H$, the server generates the staleness threshold $W_{new}$ for the current round:
\begin{equation}\label{eq:threshhold}
    \begin{cases}
        W_{new}' = W_{old} \times (1 - \lambda \times \frac{H_{new}-H_{old}}{H_{old}})\\
        W_{new} = W_{new}' \times (1 + \mu \times \frac{N_{new}-N_{old}}{N_{old}})
    \end{cases}
\end{equation}
where $N_{new}$ represents the number of devices that need to be distributed with the global model according to $W_{new}'$, $N_{old}$ is the count in the previous round. The coefficients $\lambda$ and $\mu$ are employed to balance the impact of staleness and communication cost on threshold adjustment, respectively. In this way, if the average model staleness significantly increases, the threshold is reduced to mitigate staleness. Conversely, if the communication cost becomes prohibitive, the threshold is increased to reduce the number of global model distributions.

\begin{algorithm}[t]
	\caption{One round process in FLUDE.} \label{alg:FLUDE}
    
    \SetAlgoNoEnd
    \SetAlgoNoLine
    
	\KwIn{Communication budget $B_{\max}$, penalty $\sigma$, exploration factor $\epsilon$, max round duration $T$, global model $\Omega_{g}$}
 
	\KwOut{Updated global model $\Omega_{g}$}

    $\mathbb{C}\leftarrow\emptyset${\color[RGB]{148,0,211}\footnotesize{\tcp*{Explored devices.}}}

    \SetKwFunction{Training}{RoundProcessOnServer}
    \SetKwProg{Fn}{Function}{:}{}
    \Fn{\Training{}}{
    
        $\mathbb{D}\leftarrow$ RegisterOnlineDevice()\;\label{line:online}

        $X \leftarrow |\mathbb{D}|${\color[RGB]{148,0,211}\footnotesize{\tcp*{Participants size.}}}\label{line:size}\label{line:size}

        $B_{pred.} \leftarrow B_{\max} + 1${\color[RGB]{148,0,211}\footnotesize{\tcp*{Predicted comm. cost.}}}

        \While{$B_{pred.} > B_{\max}$}{
        
            $X \leftarrow X \times \frac{B_{\max}}{B_{pred.}}$\;\label{line:predX}

            {\color[RGB]{148,0,211}\footnotesize{\tcc{Select participants based on Algorithm \ref{alg:selection}.}}}
            $\mathbb{S} \leftarrow$ SelectParticipant($\mathbb{D}$, $\mathbb{C}$, $X$, $\epsilon$, $\sigma$)\;\label{line:select}
            
            $\mathbb{S}_{distr.} \leftarrow$ StalenessAwareModelDistri($\mathbb{S}$)\;\label{line:distribute}

            $R \leftarrow$ CalculateAverageDependability($\mathbb{S}$)\;\label{line:calDependability}

            {\color[RGB]{148,0,211}\footnotesize{\tcc{Calculate predicted communication cost.}}}
            $B_{pred.} \leftarrow |\mathbb{S}_{distr.}| + |\mathbb{S}| \times R$\;\label{line:predict}
        }

        Distribute $\Omega_{g}$ to devices in $\mathbb{S}_{distr.}$\;

        {\color[RGB]{148,0,211}\footnotesize{\tcc{Wait for devices to upload local models.}}}
        $\mathbb{W} \leftarrow \emptyset${\color[RGB]{148,0,211}\footnotesize{\tcp*{Received local models.}}}\label{line:waits}

        $\Delta T \leftarrow 0${\color[RGB]{148,0,211}\footnotesize{\tcp*{Round duration.}}}

        \While{$|\mathbb{W}| < |\mathbb{S}| \times R$ and $\Delta T < T$}{
            $\Delta T$, $\mathbb{W}\leftarrow$ ReceiveOneLocalModel()\;\label{line:waite}
        }


        {\color[RGB]{148,0,211}\footnotesize{\tcc{Aggregate received local models.}}}
        $\Omega_{g} \leftarrow$ AggregateLocalModels($\mathbb{W}$)\;
        \textbf{Return} $\Omega_{g}$\;\label{line:agge}
        
    }
\end{algorithm}

\subsection{Integrating Modules into FLUDE}
Lastly, we integrate the aforementioned modules into FLUDE, as described in Algorithm \ref{alg:FLUDE}. 
Specifically, at the beginning of each round, FLUDE collects the status from devices to register the online devices available for training (Line \ref{line:online}). 
Then, the server selects $X$ online devices to participate in current training round (Line \ref{line:size}-\ref{line:predict}). 
To reduce unnecessary communication costs, the server only distributes the global model to devices whose locally cached models are overly stale, as proposed in Section \ref{subsec:distribution} (Line \ref{line:distribute}). 
Considering the communication budget $B_{\max}$, FLUDE adaptively determines the proper number $X$ of participants  (Line \ref{line:predX}) by predicting communication costs of model transmission (including both mode distribution and reception) according to the average device dependability of selected devices (Line \ref{line:calDependability}-\ref{line:predict}). 
After device selection and model distribution, the server waits for the selected devices to complete their local training and upload the updated local models. In undependable environments, it is challenging to predict the actual training time for each device. Therefore, setting a fixed round deadline will lead to considerable idle waiting time from undependable devices, which severely impacts system efficiency. To shorten the round duration, each round has two termination conditions. One is that the round's duration exceeds a time threshold $T$, the other is that the central server receives $|\mathbb{S}| \times R$ local models that arrive first, where $\mathbb{S}$ are selected devices in the current round and $R$ is the average dependability of devices in $\mathbb{S}$. Once either condition is met, the round concludes (Lines \ref{line:waits}-\ref{line:waite}). Lastly, the received local models from selected devices are aggregated to update the global model (Line \ref{line:agge}). Note that FLUDE measures the dependability and the model staleness of a device without revealing its raw data. Consequently, no privacy information (\eg, data distribution) about devices can be disclosed.

%% file: content/simulation.tex
\subsection{System Implementation} 
We implement FLUDE on two physical platforms, where we utilize an AMAX deep learning workstation with an Intel(R) Xeon(R) Platinum 8358P CPU, 8 NVIDIA RTX A6000 (49GB) GPUs and 512GB RAM as the central server.
One platform is constructed using 40 OPPO smartphones \cite{oppo}, including 10 Renos, 15 Finds, and 15 As, which represent devices with high, medium, and low computing capabilities, respectively.
Each smartphone conducts model training based on the MNN framework \cite{mnn}.
The other platform includes 80 NVIDIA Jetson devices \cite{jetson}, comprising 30 Jetson TX2s, 40 Jetson NXs, and 10 Jetson AGXs. Specifically, TX2 and NX can work in one of four computation modes while AGX has eight modes \cite{jetson}. Devices working in different modes exhibit diverse computing capabilities. 
In this platform, we build the software platform based on Docker Swarm \cite{merkel2014docker} and the PyTorch deep learning library \cite{paszke2019pytorch}. 

\subsection{Experimental Settings}\label{sec:experimental_settings}
\textbf{Device Undependability.} In practice, devices may become undependable during local training and finally fail to upload their local models to the server. To this end, we set an undependability rate for each device to simulate the probability of becoming undependable during local training (\eg, violating training conditions and experiencing software/hardware malfunctions). Specifically, we divide all devices into three groups, each representing devices with high, medium, and low dependability, respectively. According to observations of our real-world FL system, the undependability rate of devices within the same group follows a normal distribution, with the means of 0.2, 0.4, and 0.6, respectively, and the variance of 0.04 for all. During local training, each device randomly becomes undependable based on its undependability rate.\\
\textbf{Participation Dynamics.} In real-world FL systems, users may choose to kill the FL process on devices if they lose interest in services provided by the FL application. Therefore, devices in FL systems will dynamically be offline. To simulate the offline behaviors of devices, in our work, each device randomly updates its state (online or offline) every 10 minutes.
Considering that the online rate varies greatly across different devices, we assign a random online rate between 0.2 and 0.8 to each device. 
When device $i$ in the FL system needs to update its state, it will generate a random decimal between 0 and 1. 
If this decimal is less than or equal to the online rate of device $i$, then device $i$ is considered to be in an online state; Otherwise, it is deemed to be in an offline state and is unable to participate in training.\\
\textbf{Bandwidth Heterogeneity.} All devices in the two platforms are connected to the central server via WiFi routers. To emulate heterogeneous bandwidths on different devices, we divide the devices into four groups, which are bound to four routers with different bandwidths. In each group, devices are placed at different locations, \eg, 2m, 8m, 14m, and 20m away from the WiFi router. Due to random channel noise and competition among devices, the bandwidth between the central server and devices dynamically varies during the training. The bandwidth of devices fluctuates between 1Mb/s and 30Mb/s, which is measured by iperf3 \cite{iperf}.\\
\begin{table*}[t]
    \centering
    \caption{Summary of final accuracy and time-to-accuracy on the four datasets.} \label{table:end-to-end}
    \resizebox{179mm}{!}{
        \centering
        \begin{tabular}{c|c|cc|cc|cc|cc|cc}
            \toprule
             \multirow{2}*{Dataset} & \multirow{2}*{Target ACC./AUC} & \multicolumn{2}{c|}{AsyncFedED} & \multicolumn{2}{c|}{SAFA} & \multicolumn{2}{c|}{FedSEA} & \multicolumn{2}{c|}{Oort} & \multicolumn{2}{c}{FLUDE} \\
             & & ACC./AUC & Time & ACC./AUC & Time & ACC./AUC & Time & ACC./AUC & Time & ACC./AUC & Time \\
            \midrule
            CIFAR-10 & 82\% & 82.68\% & 8.12h & 83.57\% & 7.23h & 85.67\% & 4.41h & 86.41\% & 3.57h & \textbf{87.52\%} & \textbf{2.68h}\\
            \midrule
            CIFAR-100 & 67\% & 67.61\% & 22.53h & 67.97\% & 19.29h & 69.92\% & 18.15h & 72.02\% & 12.93h & \textbf{73.31\%} & \textbf{9.39h}\\
            \midrule
            Google Speech & 78\% & 78.55\% & 5.47h & 79.19\% & 5.04h & 82.14\% & 3.07h & 81.21\% & 2.25h & \textbf{82.44\%} & \textbf{1.37h}\\
            \midrule
            Avazu & 79\% & 80.03\% & 3.18h & 79.57\% & 3.51h & 81.83\% & 2.57h & 81.05\% & 3.12h & \textbf{83.38\%} & \textbf{2.13h}\\
            \bottomrule
        \end{tabular}
    }
\end{table*}
\begin{figure*}[t]\centering
    \begin{minipage}[t]{0.246\linewidth}\centering
        \subfigure[CIFAR-10]{\centering
                \label{fig:time-to-accuracy-end-to-end-cifar10}
                \includegraphics[width=1.0\linewidth]{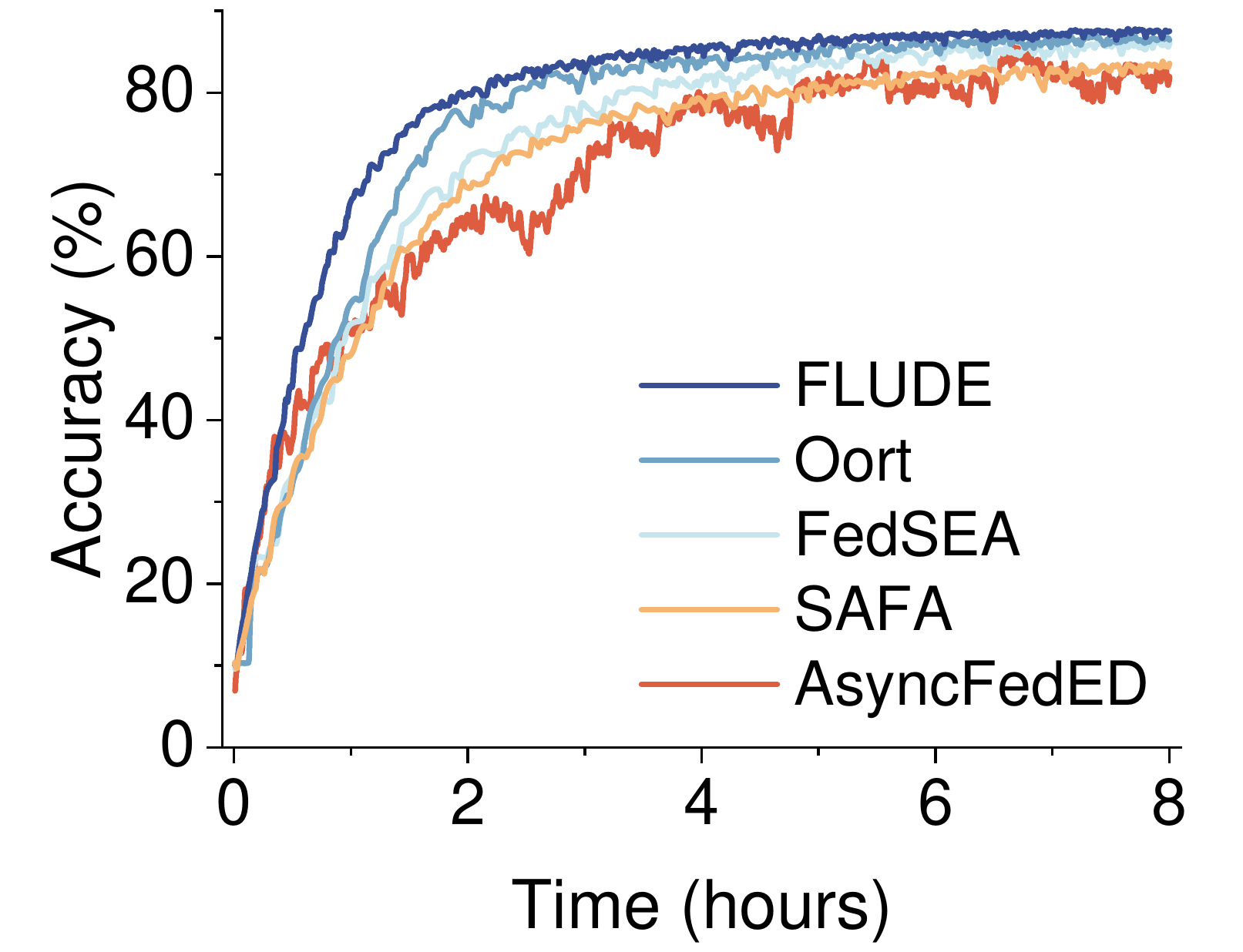}
            }
    \end{minipage}
    \begin{minipage}[t]{0.246\linewidth}\centering
        \subfigure[CIFAR-100]{\centering
                \label{fig:time-to-accuracy-end-to-end-cifar100}
                \includegraphics[width=1.0\linewidth]{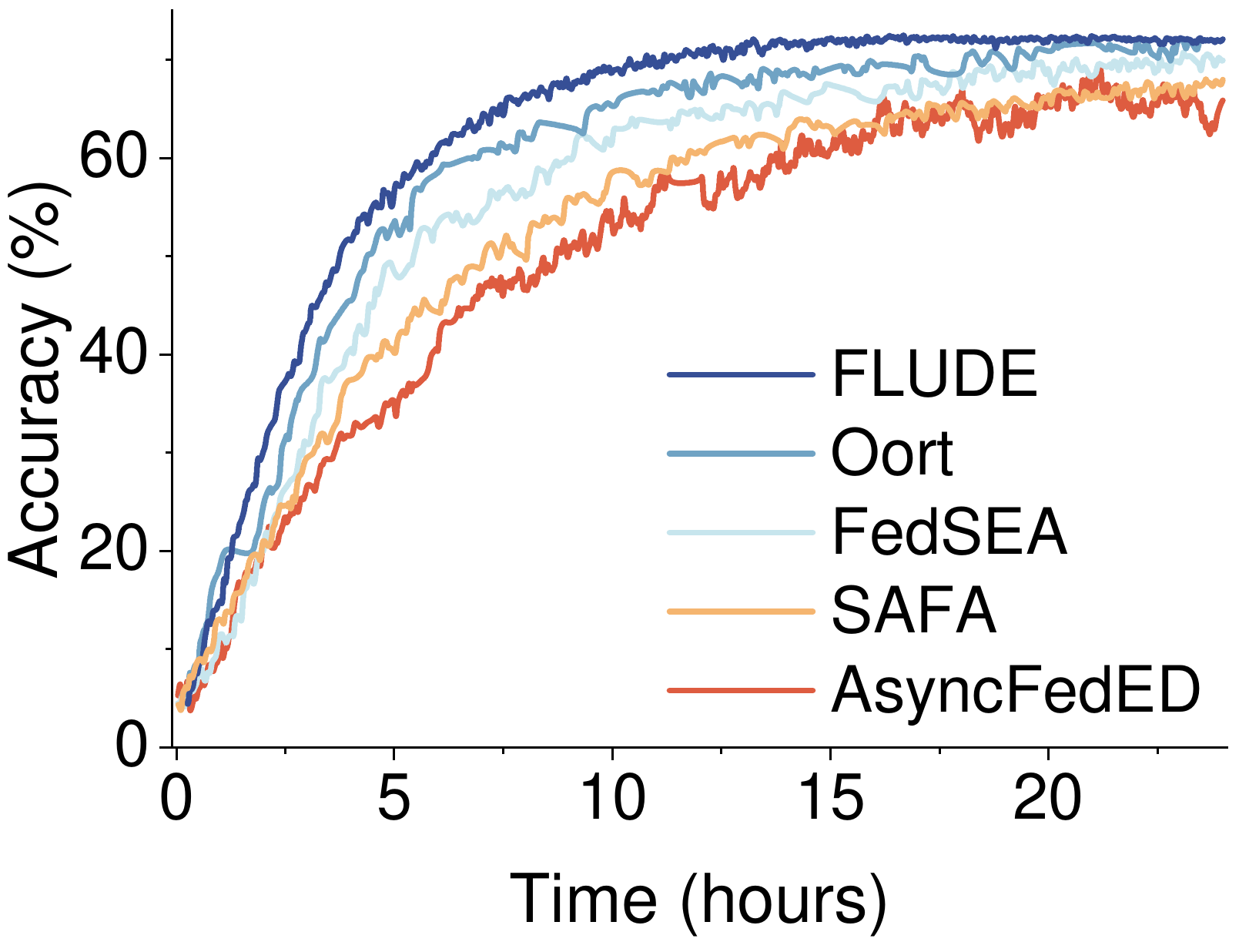}
            }
    \end{minipage}
    \begin{minipage}[t]{0.246\linewidth}\centering
        \subfigure[Google Speech]{\centering
                \label{fig:time-to-accuracy-end-to-end-googlespeech}
                \includegraphics[width=1.0\linewidth]{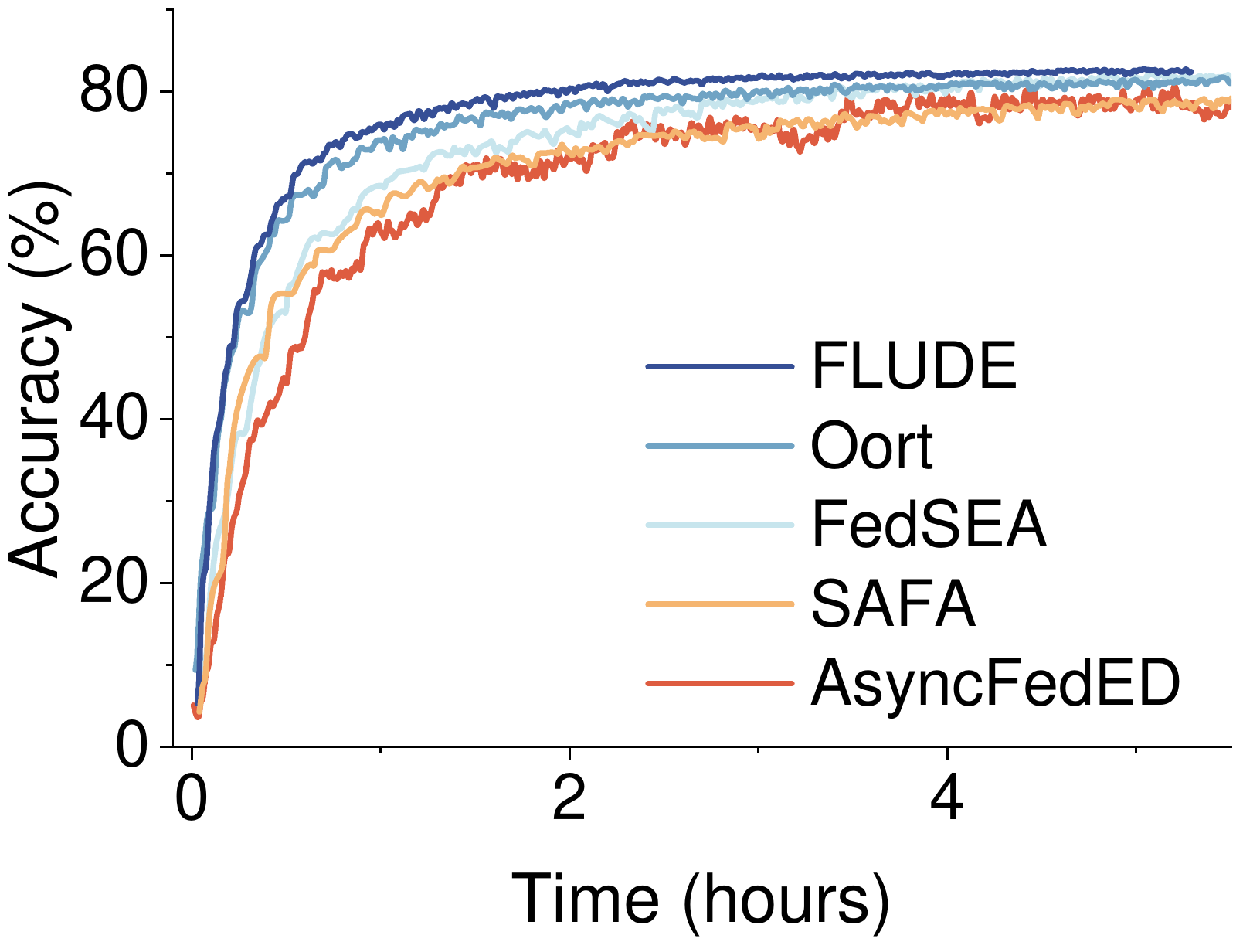}
            }
    \end{minipage}
    \begin{minipage}[t]{0.246\linewidth}\centering
        \subfigure[Avazu]{\centering
                \label{fig:time-to-accuracy-end-to-end-avazu}
                \includegraphics[width=1.0\linewidth]{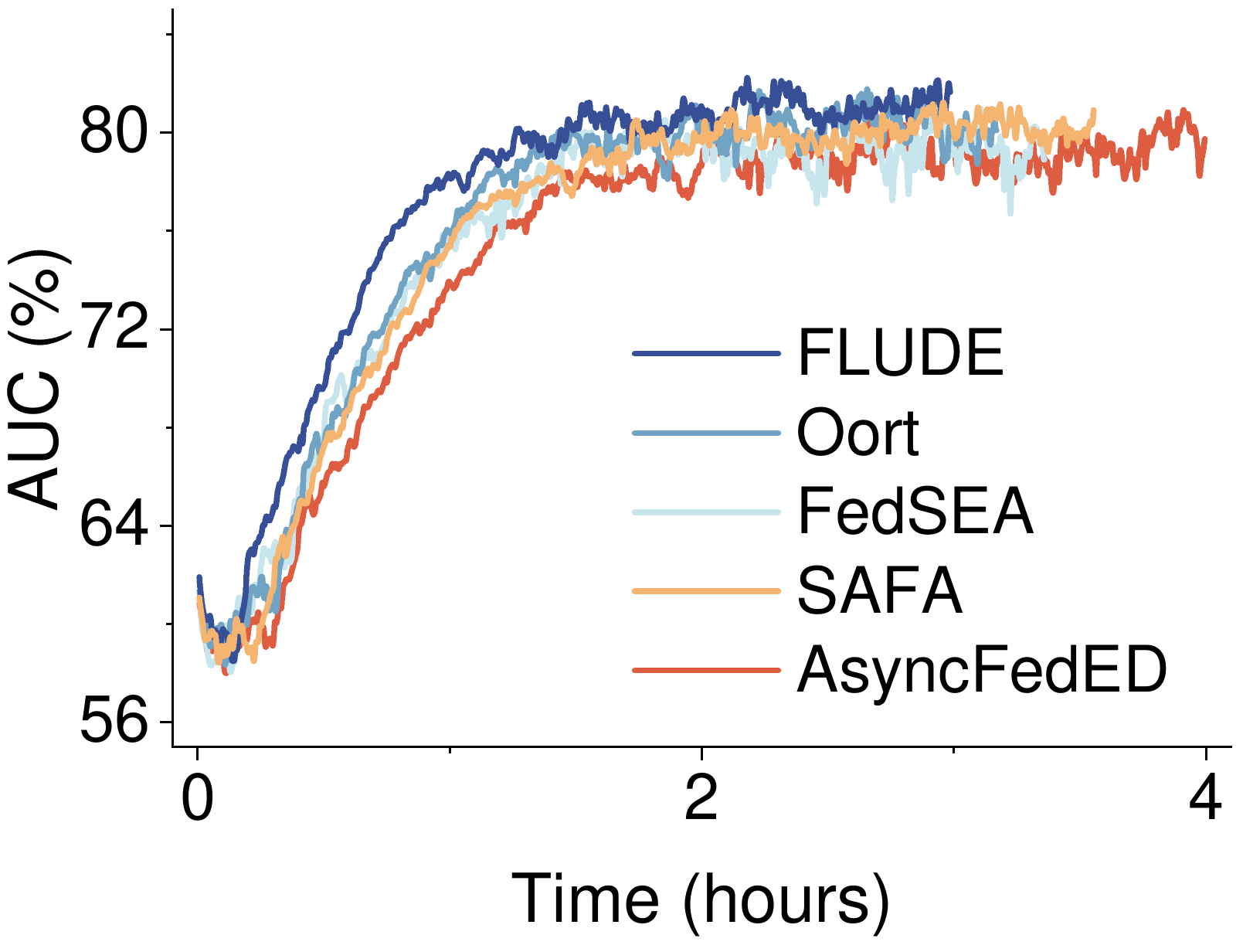}
            }
    \end{minipage}
    \caption{Performance comparison of time-to-accuracy between FLUDE and the baselines.}\label{fig:time-to-accuracy-end-to-end}
\end{figure*}
\textbf{Tasks, Datasets, and Models.} To demonstrate the generality of FLUDE across different tasks, datasets, and models, we evaluate FLUDE on four popular real-world datasets, which cover three common task scenarios, \ie, image classification, speech recognition, and federated recommendation.
\begin{itemize}[leftmargin=*]
    \item \textbf{Image Classification \cite{rawat2017deep}} is a classic task in computer vision and has been widely applied to mobile devices. We train two well-known models on two real-world datasets, \ie, VGG-9 \cite{vgg9} on CIFAR-10 \cite{krizhevsky2009learning} and ResNet-18 \cite{he2016deep} on CIFAR-100 \cite{krizhevsky2009learning}, for evaluation. Specifically, both CIFAR-10 and CIFAR-100 contain 60,000 images, with 50,000 used for training and 10,000 for testing. 
    We divide the training data into 80 parts and assign them to 80 Jetson devices. Images in CIFAR-10 can be categorized into 10 classes, with each device randomly holding 4-class data, mimicking non-IID data in practical FL \cite{li2021hermes}. For CIFAR-100 with 100 classes, we randomly assign 40 classes to each device for training.
    \item \textbf{Speech Recognition \cite{lee2021biosignal}} is widely deployed on smartphones for voice assistants in recognizing human voices. In this task, we train a convolutional neural network with four 1-d convolution layers \cite{cnn} on the Google speech dataset \cite{warden2018speech}, which contains audio clips of 35 target words (such as "yes", "bed", \etc), totaling 85,511 training audio clips and 4,890 testing audio clips. We divide the training dataset into 80 parts, each containing random audio clips of 10 target words, serving as the training data on each Jetson device.
    \item \textbf{Federated Recommendation \cite{yin2024device}} is widely applied in e-commerce, advertising, and news platforms.
    In this task, we train a well-known model, \ie, WideAndDeep \cite{cheng2016wide}, on the Avazu dataset \cite{avazu} for online advertisement click-through rate (CTR) prediction.
    In Avazu, each record represents a user's click behavior on an ad displayed on a mobile device. We divide the original dataset according to the \emph{deviceID} feature domain, with each part representing the data of a device. 
    Since the number of data parts is larger than that of smartphones, we assign several data parts to each smartphone. We divide each smartphone's data into training and test sets with a ratio of 8:2.
\end{itemize}
\textbf{Baselines.} To comprehensively evaluate the performance of FLUDE, we choose the following four baselines:
\begin{itemize}[leftmargin=*]
    \item \textbf{Oort \cite{lai2021oort}} is a classic synchronous FL system with efficient device selection, considering data and system heterogeneity simultaneously. 
    Specifically, Oort selects a subset of devices with strong computation and communication capabilities as well as high-quality data to participate in training in each round.
    \item \textbf{AsyncFedED \cite{wang2022asyncfeded}} is an asynchronous FL framework with model aggregation optimization, which measures the staleness of each local model by considering the Euclidean distance between the local model and the current global model. The optimized aggregation weights for models are then determined based on the staleness of the models.
    \item \textbf{SAFA \cite{wu2020safa}} is a semi-asynchronous FL framework to address problems such as idle waiting time on servers and poor convergence rate in FL. In SAFA, a part of the devices need to be synchronized with the server in each round.
    \item \textbf{FedSEA \cite{sun2022fedsea}} is also a semi-asynchronous FL system. Different from SAFA, FedSEA balances the local training time of devices by reducing the number of local training iterations of devices with slow training speeds. 
\end{itemize}
\textbf{Evaluation Metrics.} We use three metrics to evaluate the performance of FLUDE and the baselines.
\begin{itemize}[leftmargin=*]
    \item \textbf{Time-to-Accuracy} \cite{li2022pyramidfl} is defined as the wall clock time for training a model to reach a target accuracy. For simplicity, we set the target accuracy to be the minimum achievable test accuracy among FLUDE and all the baselines for fair comparison.
    \item \textbf{Final Accuracy (ACC.)/Area Under Curve (AUC)} \cite{li2022pyramidfl} are defined as the accuracy/area under curve on the test datasets obtained by the trained model. We adopt the ACC. for the Image Classification and Speech Recognition tasks, and the AUC for the Federated Recommendation task.
    \item \textbf{Communication Cost} is defined as the network traffic consumed for transmitting model parameters between devices and the server (including both uploading and downloading) to reach a target accuracy.
\end{itemize}
\textbf{Parameter settings.}
The mini-batch size is 32 for all tasks. The initial learning
rate is 0.04 for CIFAR-10, 0.1 for CIFAR-100 and Avazu, and 0.01 for Google Speech. In configuring the training selector, FLUDE uses the popular exploration factor \cite{auer2002finite}, where the initial exploration factor is 0.9, and decreases by a factor of 0.98 after 
each round when it is larger than 0.2. 
We set the penalty $\sigma$ of participation frequency to 0.5, coefficients $\lambda$ and $\mu$ in Eq. \eqref{eq:threshhold} to 1 and 0.5, respectively. 
\begin{figure*}[t]\centering
    \begin{minipage}[t]{0.246\linewidth}\centering
        \subfigure[CIFAR-10]{\centering
                \label{fig:communication-cost-end-to-end-cifar10}
                \includegraphics[width=1.0\linewidth]{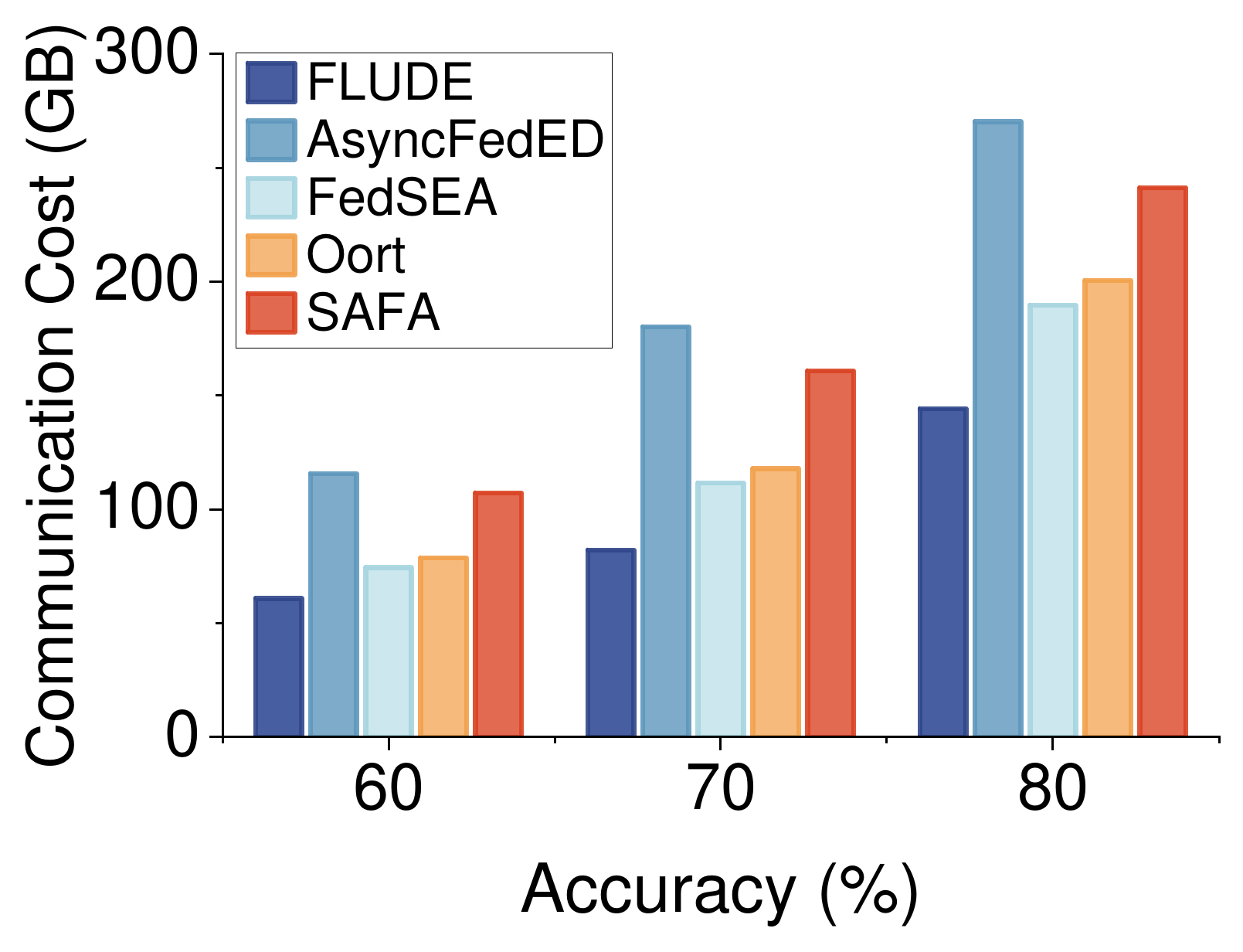}
            }
    \end{minipage}
    \begin{minipage}[t]{0.246\linewidth}\centering
        \subfigure[CIFAR-100]{\centering
                \label{fig:communication-cost-end-to-end-cifar100}
                \includegraphics[width=1.0\linewidth]{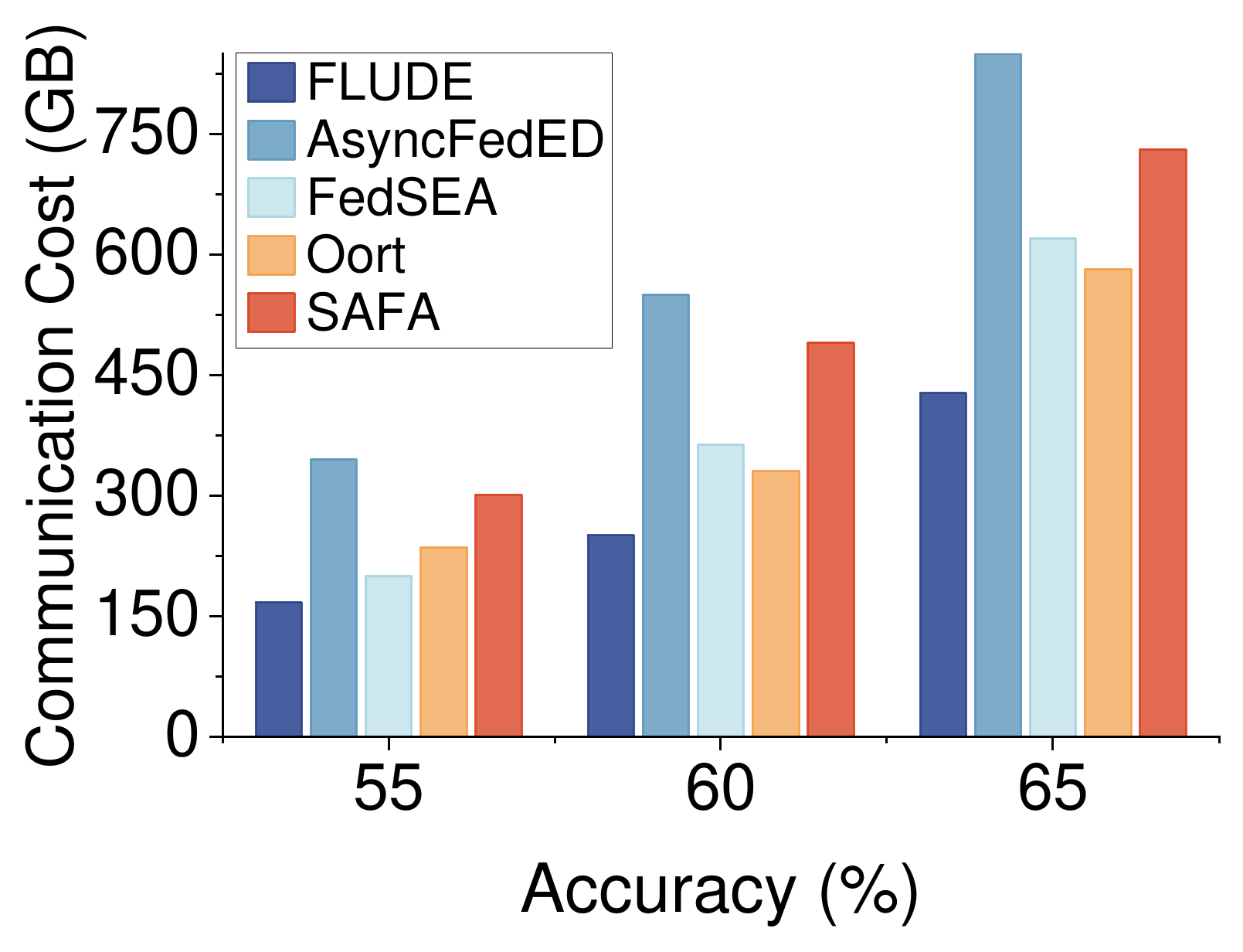}
            }
    \end{minipage}
    \begin{minipage}[t]{0.246\linewidth}\centering
        \subfigure[Google Speech]{\centering
                \label{fig:communication-cost-end-to-end-googlespeech}
                \includegraphics[width=1.0\linewidth]{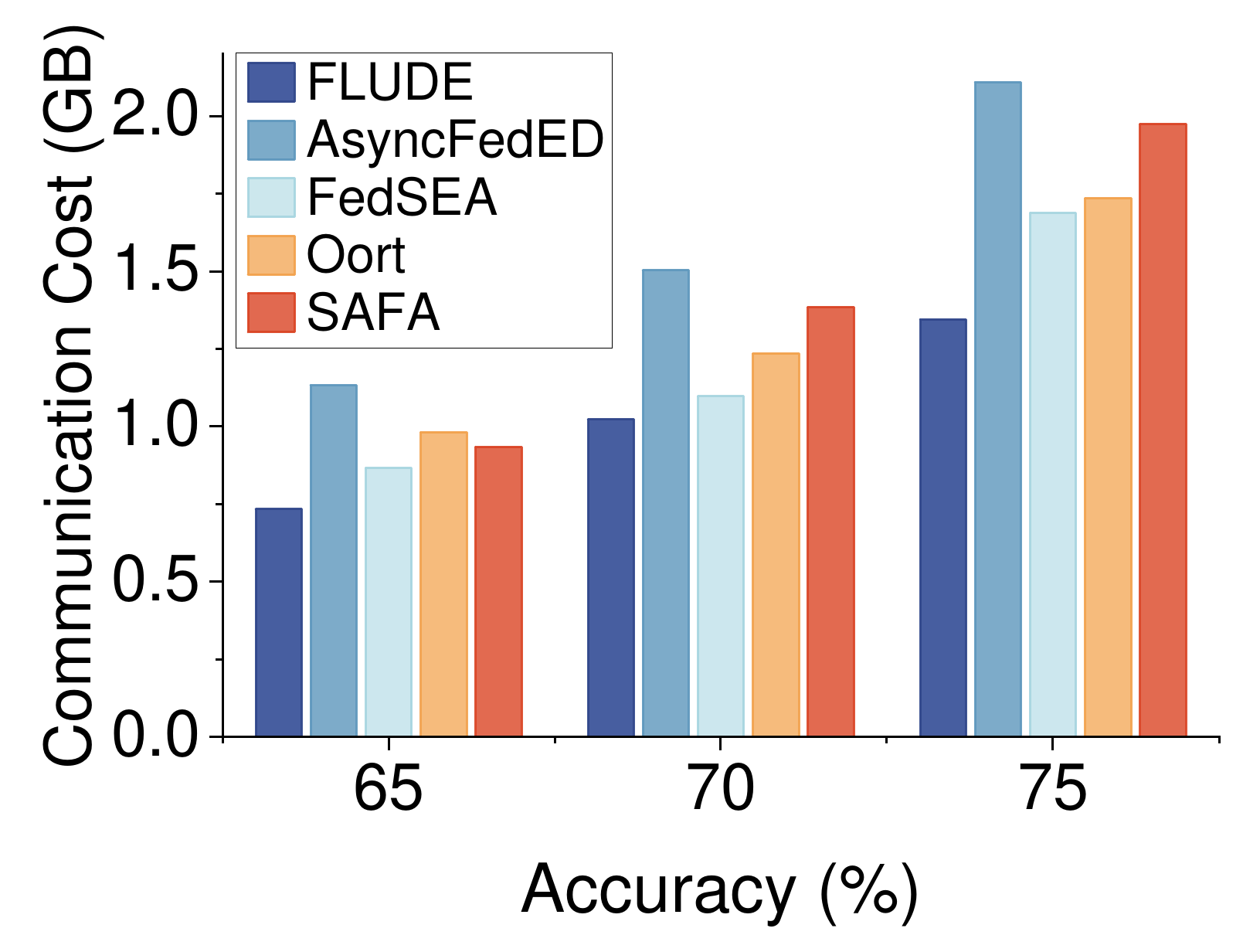}
            }
    \end{minipage}
    \begin{minipage}[t]{0.246\linewidth}\centering
        \subfigure[Avazu]{\centering
                \label{fig:communication-cost-end-to-end-avazu}
                \includegraphics[width=1.0\linewidth]{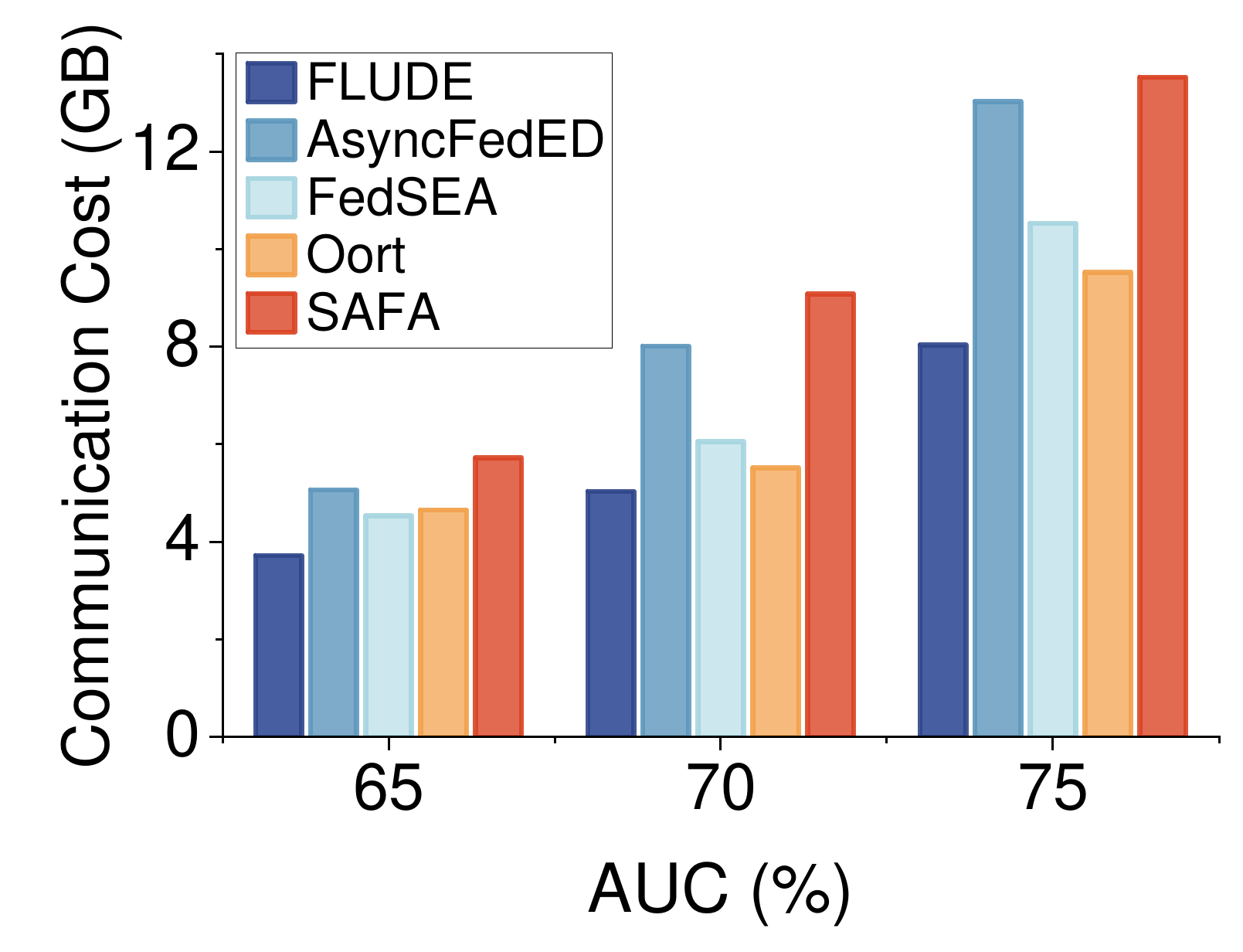}
            }
    \end{minipage}
    \caption{Comparison of communication costs between FLUDE and the baselines.}\label{fig:communication-cost-end-to-end}
\end{figure*}
\subsection{Overall Performance}
We begin by comparing the overall performance of FLUDE with the baselines on all four datasets.\\
\textbf{FLUDE speeds up the model training to reach the final target accuracy.} 
In Table \ref{table:end-to-end}, given the Google speech dataset in the speech recognition task, FLUDE reaches 1.6$\times-$4.1$\times$ faster than the baselines in terms of required wall clock time to achieve the same final target accuracy of 78\%. 
When tested with CIFAR-100, FLUDE demonstrates a speedup of 1.4$\times$ to 2.4$\times$. Figure \ref{fig:time-to-accuracy-end-to-end} also shows the consistent speedup of FLUDE, which remains unaffected by various models and datasets. For instance, by Figure \ref{fig:time-to-accuracy-end-to-end-cifar10}, to achieve the target accuracy of 82\% on CIFAR-10, the required wall clock time of FLUDE is 2.68 hours, while AsyncFedED, SAFA, FedSEA, and Oort require 8.12, 7.23, 4.41, and 3.57 hours, respectively. This results in a speedup for FLUDE ranging from 1.3$\times$ to 3.0$\times$ on CIFAR-10.
In Figure \ref{fig:time-to-accuracy-end-to-end-avazu}, FLUDE achieves speedups of 1.2$\times-$1.6$\times$ compared to the baselines with the target accuracy of 79\% on Avazu.
These speedups with the wall clock time reduction stem from selecting devices with high dependability in each round and caching the models on undependable devices, which allows more devices to successfully complete training on time and reduce idle waiting time for undependable devices on the central server. Besides, FLUDE can reach the final target accuracy with fewer training rounds than baselines due to the balanced participation frequency among devices and reduced model staleness.\\
\textbf{FLUDE improves the final accuracy.}
As shown in Table \ref{table:end-to-end}, FLUDE achieves 5.70\%, 5.34\%, 3.39\%, and 1.29\% higher final accuracy compared with AsyncFedED, SAFA, FedSEA and Oort on the CIFAR-100 dataset, respectively.
The final accuracy improvement can be attributed to aggregating more local updates, \ie, incorporating more knowledge from devices' local data. By selecting more dependable devices to participate in training, FLUDE reduces the overall undependability rate, thereby enhancing final accuracy.
Besides, in undependable environments, there is a significant difference in the participation frequencies among devices, which biases the global model towards devices with higher frequencies, resulting in poor final accuracy on devices with lower frequencies. 
FLUDE effectively balances the training frequencies among devices, aligning the local data distribution on participating devices more closely with the global distribution. This balance improves the final accuracy of the global model. 
As a result, FLUDE achieves higher accuracy from 1.11\% to 5.70\% than baselines for all datasets in Table \ref{table:end-to-end}.\\
\textbf{FLUDE reduces the communication cost to reach the target accuracy.} Compared to the baselines, FLUDE can achieve lower communication costs when reaching the final target accuracy on all datasets. 
As shown in Figure \ref{fig:communication-cost-end-to-end-cifar10}, given the target accuracy of 80\% on CIFAR-10, FLUDE consumes only 145GB of communication cost while AsyncFedED, SAFA, FedSEA and Oort consume 270GB, 241GB, 190GB and 201GB, respectively. 
This reduction in communication cost is achieved by maintaining a local model cache on each device in FLUDE. When a dependable device becomes undependable, the local model is cached locally, without redistributing the global model. Additionally, FLUDE consistently selects dependable devices to participate in training, significantly decreasing the waste of communication bandwidth.
Ultimately, as shown in Figure \ref{fig:communication-cost-end-to-end}, when achieving the same target accuracy on the four datasets, FLUDE reduces the communication cost by 23.71\%-49.7\% compared to the baselines.

\subsection{Component-wise Evaluation}
In this section, we implement three breakdown versions of FLUDE to evaluate and understand the effectiveness of the key components incorporated in FLUDE.
\begin{itemize}[leftmargin=*]
    \item \textbf{FLUDE w/o device selector.} We disable the device selector, in which the central server randomly selects a set of devices to participate in training in each round. Consequently, many undependable devices may fail to upload their local models to the central server, leading to wasted communication and computation resources due to repeated model redistributions. This approach also restricts model accuracy, as it becomes biased towards devices with lower participation frequencies.
    \item \textbf{FLUDE w/o model distributor (full model distribution).} We disable the adaptive model distribution, causing the central server to distribute the latest global model to all selected devices in each round, regardless of whether they have a local model cached. This method results in excessive use of valuable communication bandwidth to distribute models to undependable devices.
    \item \textbf{FLUDE w/o model distributor (least model distribution).} In contrast to full model distribution, the least model distribution sends the latest global model only to devices whose local model cache is empty, allowing other devices to continue training based on their locally cached models. This method can lead to stale local models on some undependable devices, which negatively impacts the final accuracy of the global model.
\end{itemize}
\begin{table}[t]
    \centering
    \caption{Impacts of the device selector on FLUDE in undependable environments.} \label{table:component-wise}
    \resizebox{80mm}{!}{
        \centering
        \begin{tabular}{c|cc|cc}
            \toprule
             \multirow{2}*{Dataset} & \multicolumn{2}{c|}{FLUDE w/o device selector} & \multicolumn{2}{c}{FLUDE} \\
             & ACC./AUC. & Time & ACC./AUC & Time\\
            \midrule
            CIFAR-10 & 83.21\% & 7.98h & \textbf{87.52\%} & \textbf{2.39h}\\
            \midrule
            CIFAR-100 & 67.37\% & 15.31h & \textbf{73.31}\% & \textbf{9.01h}\\
            \midrule
            Google Speech & 78.25\% & 5.31h & \textbf{82.44\%} & \textbf{1.27h}\\
            \midrule
            Avazu & 79.01\% & 4.31h & \textbf{83.38\%} & \textbf{2.03h}\\
            \bottomrule
        \end{tabular}
    }
\end{table}
\begin{figure}[t]\centering
    \begin{minipage}[t]{0.495\linewidth}\centering
        \subfigure[CIFAR-10]{\centering
                \label{component-wise-time-to-accuracy-cifar10}
                \includegraphics[width=1.0\linewidth]{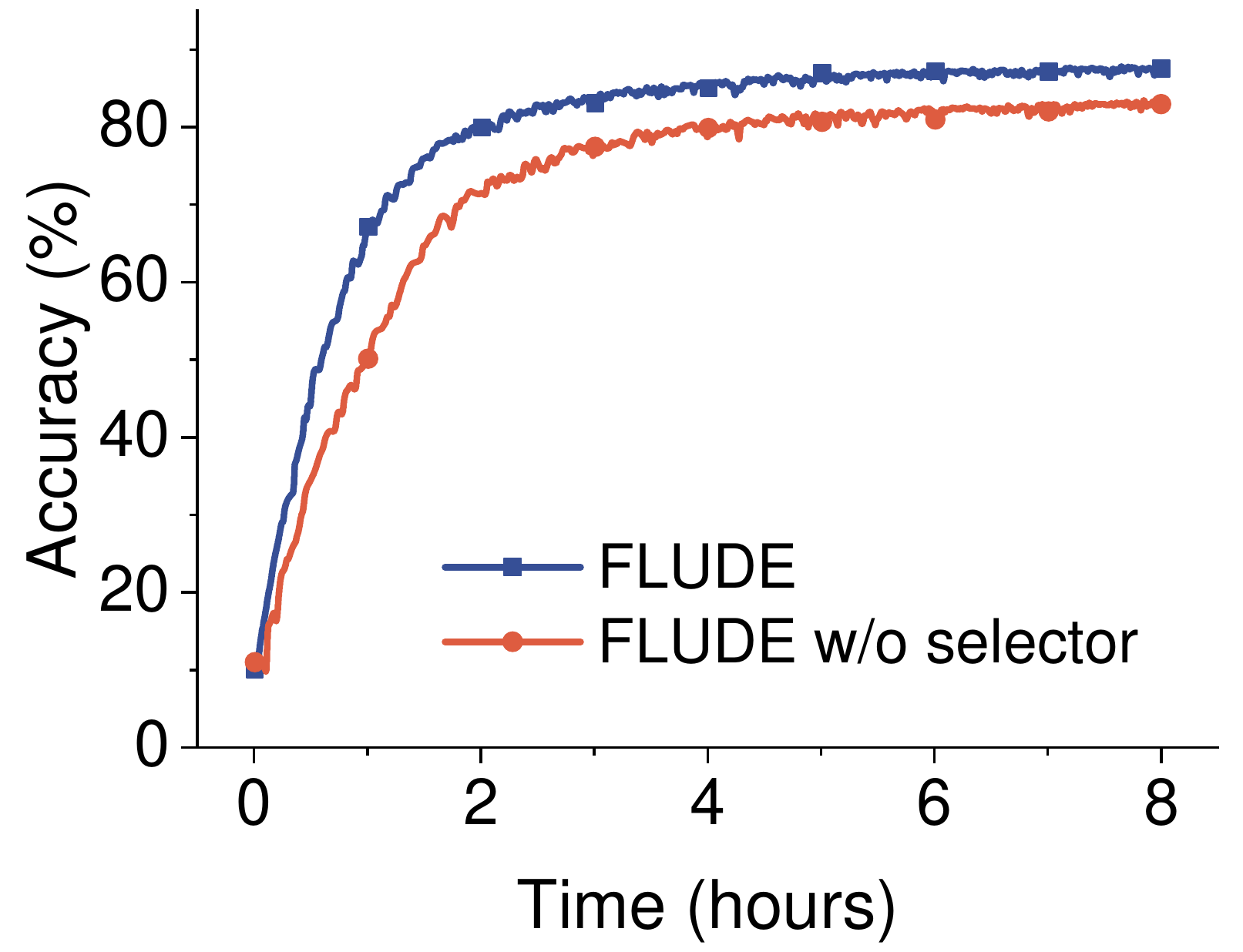}
            }
    \end{minipage}
    \begin{minipage}[t]{0.495\linewidth}\centering
        \subfigure[Google Speech]{\centering
                \label{component-wise-time-to-accuracy-googlespeech}
                \includegraphics[width=1.0\linewidth]{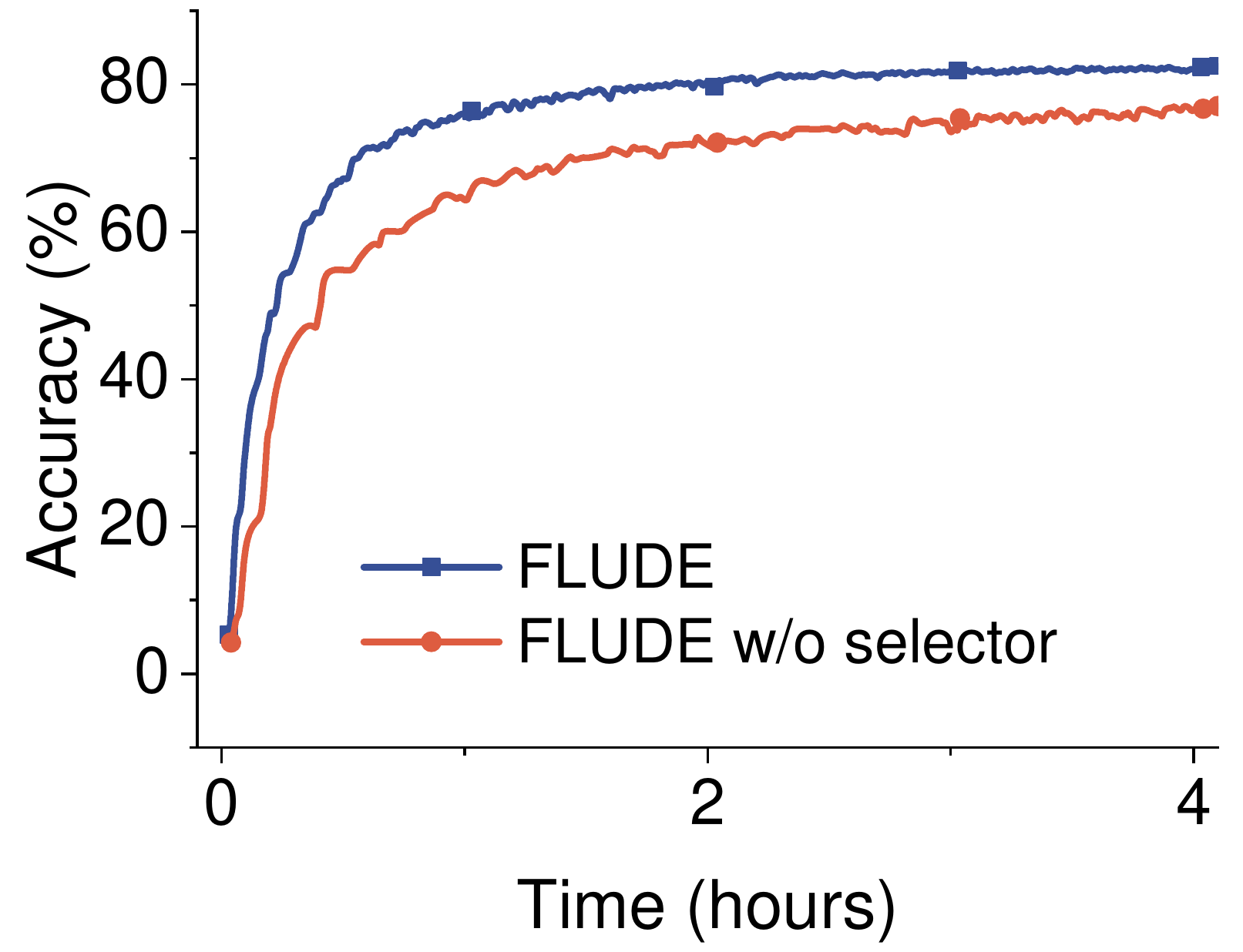}
            }
    \end{minipage}
    \caption{Impact of device selector in FLUDE on time-to-accuracy.}\label{component-wise-time-to-accuracy}
\end{figure}
\textbf{FLUDE guarantees system performance and balances participation frequencies among devices via the adaptive device selection.} Without adaptive device selection, Table \ref{table:component-wise} shows that the final accuracy and training speed of FLUDE decreases significantly on all datasets. For example, given the CIFAR-100 dataset, FLUDE w/o device selector achieves 5.94\% lower final accuracy than FLUDE with the device selector. Besides, the required wall clock time for training of FLUDE w/o device selector to reach the target accuracy (67\%) increases by 69.92\%. 
We also plot the time-to-accuracy on the CIFAR-10 and Google Speech datasets in Figure \ref{component-wise-time-to-accuracy}. FLUDE w/o device selector performs substantially worse than the native FLUDE in both final accuracy and time-to-accuracy metrics.\\ 
\textbf{Remark.} The adaptive device selection significantly contributes to the improvements of FLUDE. By selecting more dependable devices for training, FLUDE ensures more robust training sessions and reduces idle waiting time for undependable devices on the central server. Additionally, balancing participation frequencies among undependable devices enhances model accuracy.\\
\begin{figure}[t]\centering
    \begin{minipage}[t]{0.48\linewidth}\centering
        \subfigure[CIFAR-100]{\centering
                \label{component-wise-accuracy-communicationcost-cifar100}
                \includegraphics[width=1.0\linewidth]{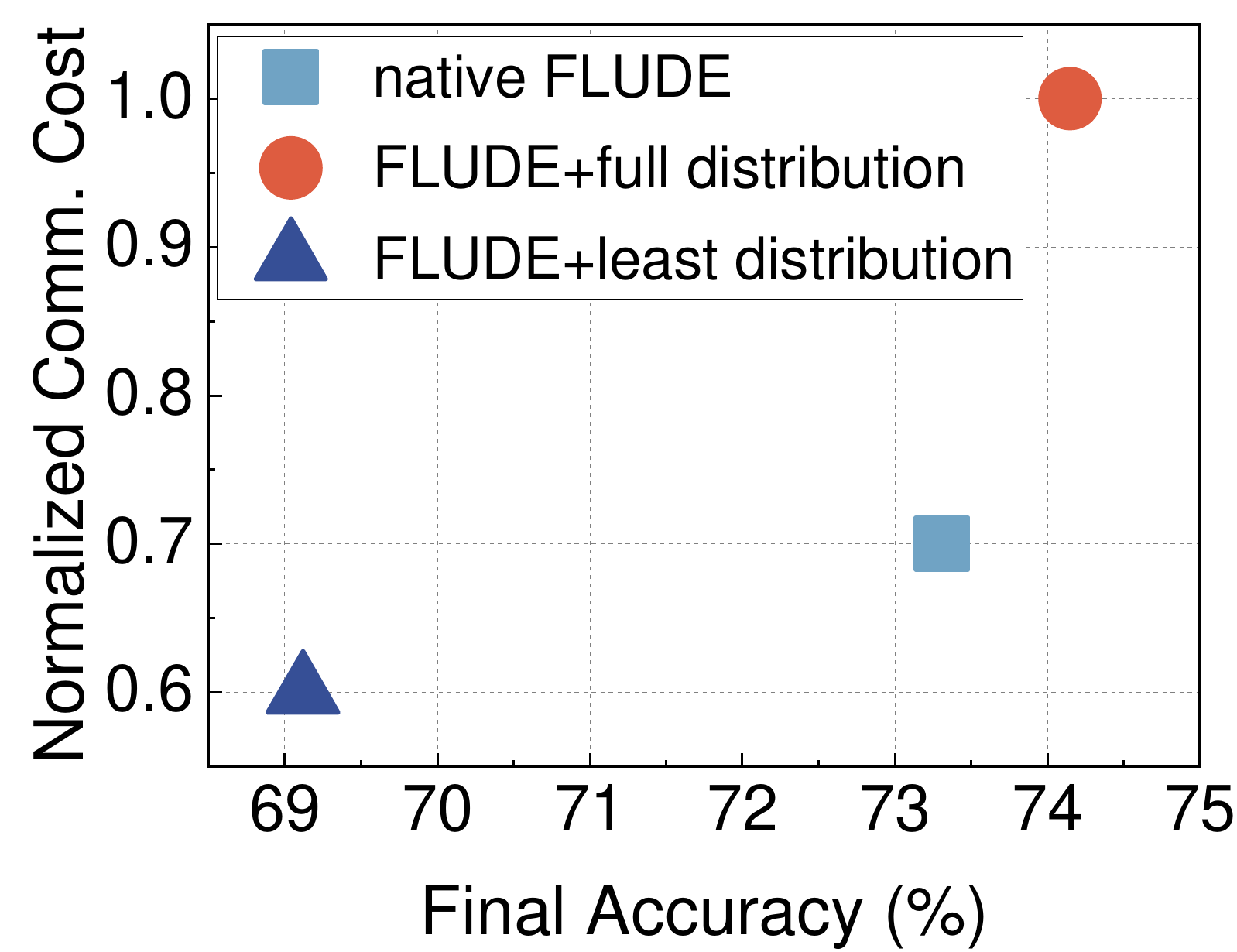}
            }
    \end{minipage}
    \begin{minipage}[t]{0.48\linewidth}\centering
        \subfigure[Google Speech]{\centering
                \label{component-wise-accuracy-communicationcost-googlespeech}
                \includegraphics[width=1.0\linewidth]{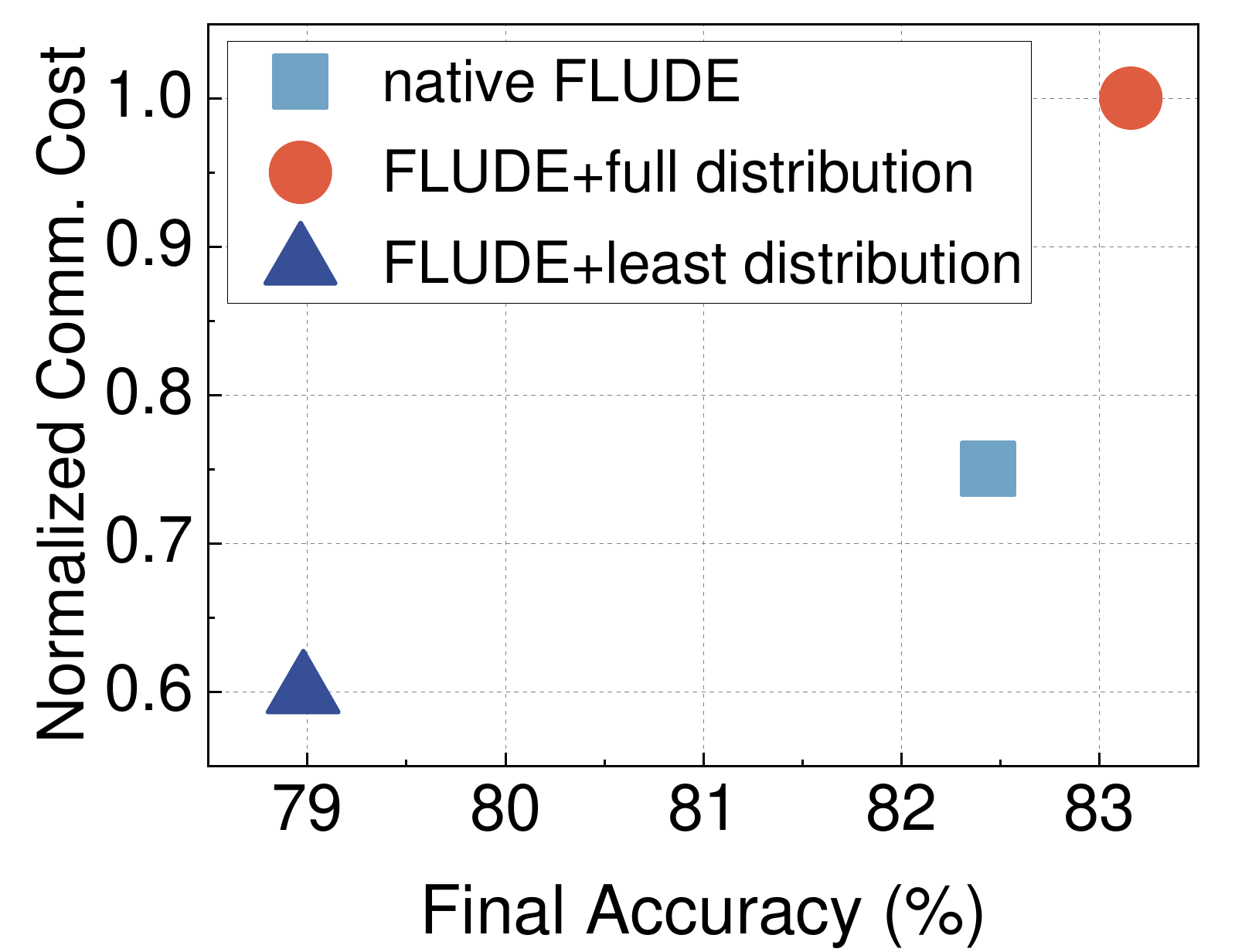}
            }
    \end{minipage}
    \caption{Impact of model distributor in FLUDE on final accuracy and communication cost.}\label{component-wise-accuracy-communicationcost}
\end{figure}
\textbf{FLUDE optimizes the communication efficiency while reducing the model staleness via adaptive model distribution.} Figure \ref{component-wise-accuracy-communicationcost} evaluates the impact of the model distributor in FLUDE on the accuracy-communication cost trade-off. Compared to FLUDE with full model distribution, FLUDE with adaptive model distribution (native FLUDE) significantly reduces communication costs. 
Specifically, adaptive model distribution helps to reduce 30\% and 25\% communication costs on the CIFAR-100 and Google Speech datasets, respectively, with only 0.76\% and 0.72\% decreases in the final accuracy. FLUDE with the least model distribution further reduces communication costs compared to native FLUDE on both datasets. However, native FLUDE significantly outperforms the least model distribution in terms of final accuracy, achieving 4.19\% and 3.46\% higher accuracy on the CIFAR-100 and Google Speech datasets, respectively.\\
\textbf{Remark.} While native FLUDE may not always surpass both full and least model distribution in final accuracy and communication cost simultaneously, it offers the best trade-off between them. 
By adaptively distributing the latest global model to a subset of devices with significantly stale local models each round, FLUDE effectively reduces the communication cost of model distribution while mitigating the negative impact of model staleness on global test accuracy.

In summary, by optimizing the device selection and model distribution, FLUDE can accelerate the training process, achieve high accuracy, and reduce communication costs simultaneously in undependable environments.

\subsection{Robustness Analysis}
We analyze FLUDE’s robustness to device undependability by answering the following question: 
\emph{Can FLUDE maintain high accuracy robustly in highly undependable environments?}
\begin{figure}[t]\centering
    \begin{minipage}[t]{0.49\linewidth}\centering
        \subfigure[CIFAR-100]{\centering
                \label{robustness-dynamics-accuracy-cifar100}
                \includegraphics[width=1.0\linewidth]{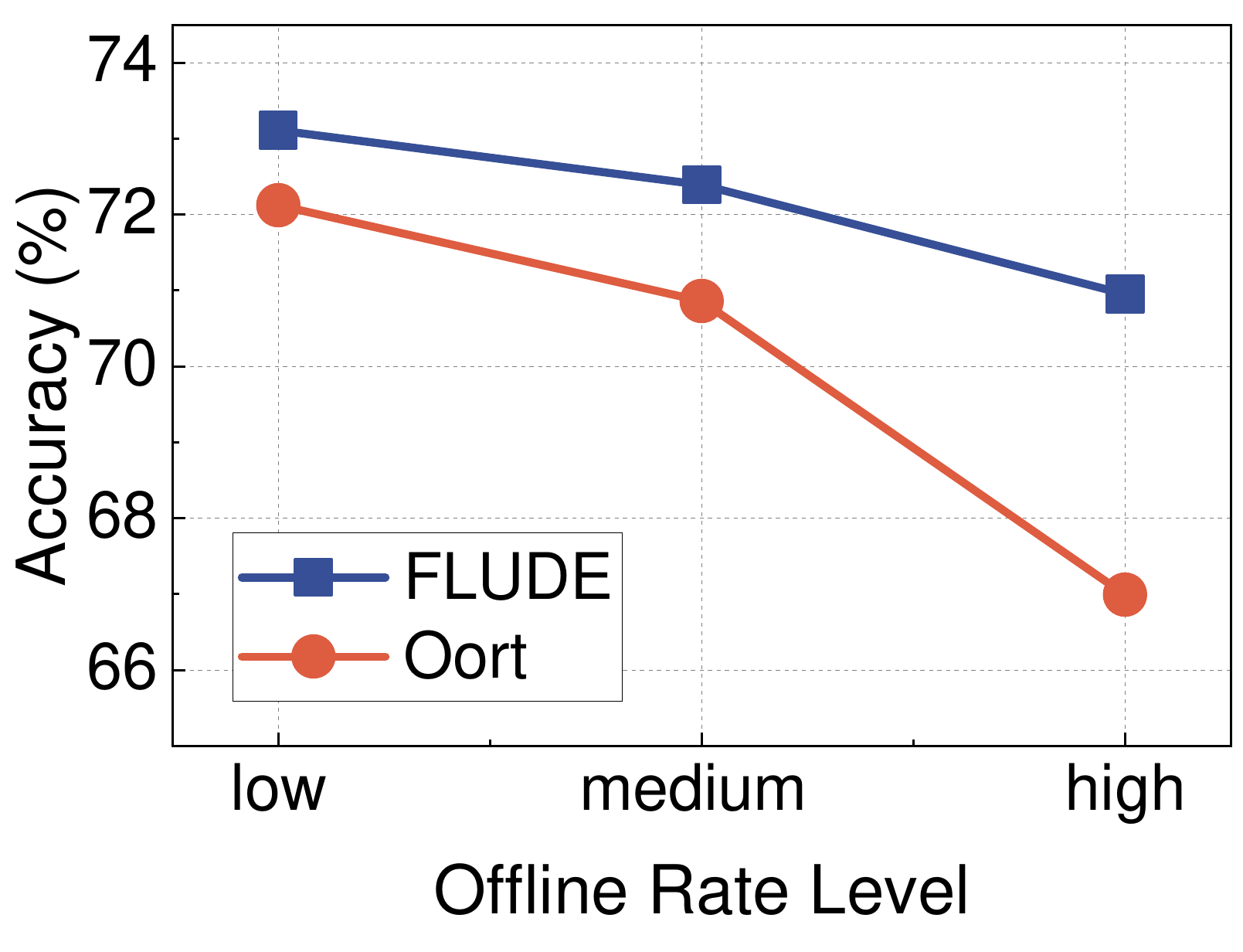}
            }
    \end{minipage}
    \begin{minipage}[t]{0.49\linewidth}\centering
        \subfigure[Google Speech]{\centering
                \label{robustness-dynamics-accuracy-googlespeech}
                \includegraphics[width=1.0\linewidth]{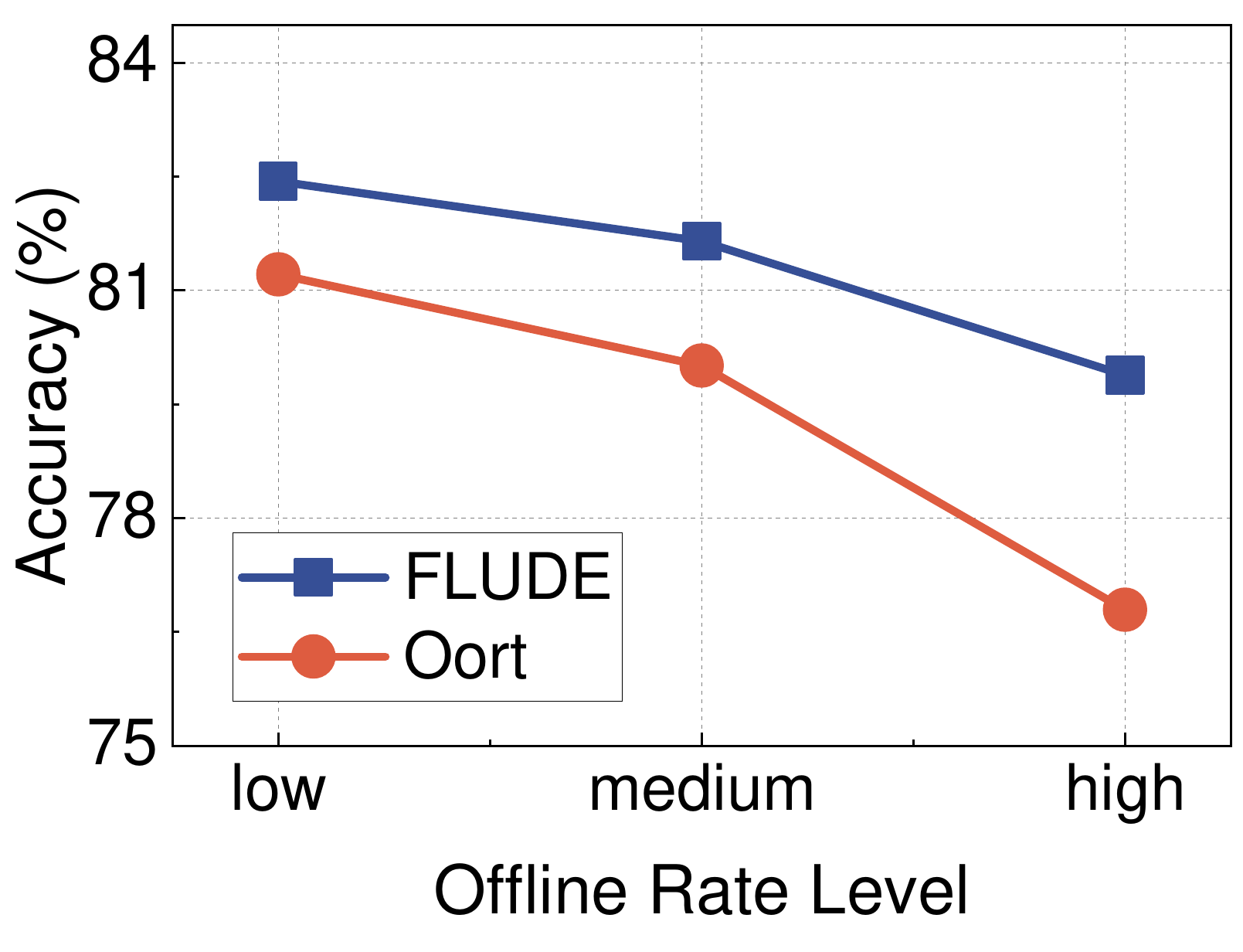}
            }
    \end{minipage}
    \caption{Impact of device offline rate on final accuracy.}\label{robustness-dynamics-accuracy}
\end{figure}
\begin{figure}[t]\centering
    \begin{minipage}[t]{0.49\linewidth}\centering
        \subfigure[CIFAR-100]{\centering
                \label{robustness-undependability-accuracy-cifar100}
                \includegraphics[width=1.0\linewidth]{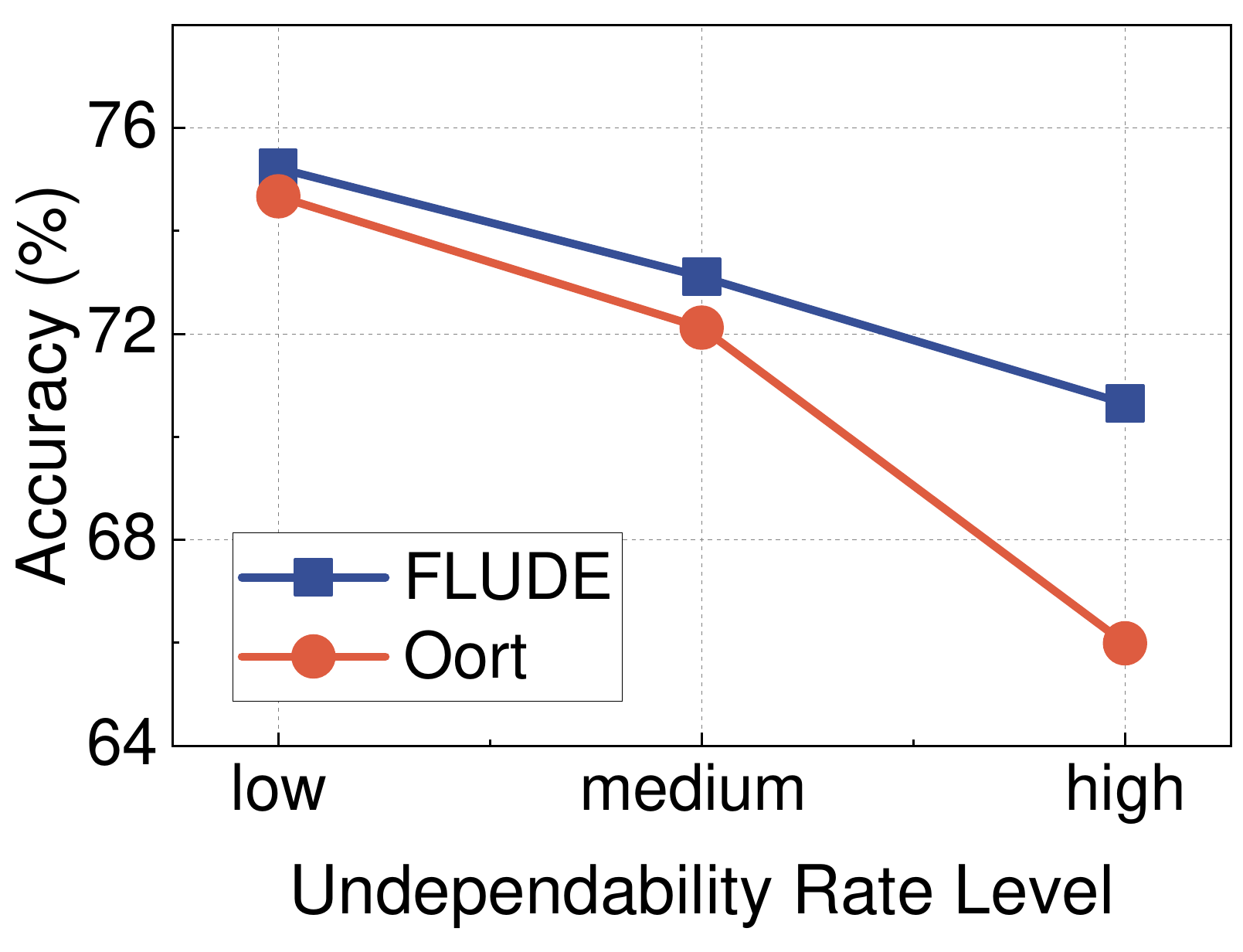}
            }
    \end{minipage}
    \begin{minipage}[t]{0.49\linewidth}\centering
        \subfigure[Google Speech]{\centering
                \label{robustness-undependability-accuracy-googlespeech}
                \includegraphics[width=1.0\linewidth]{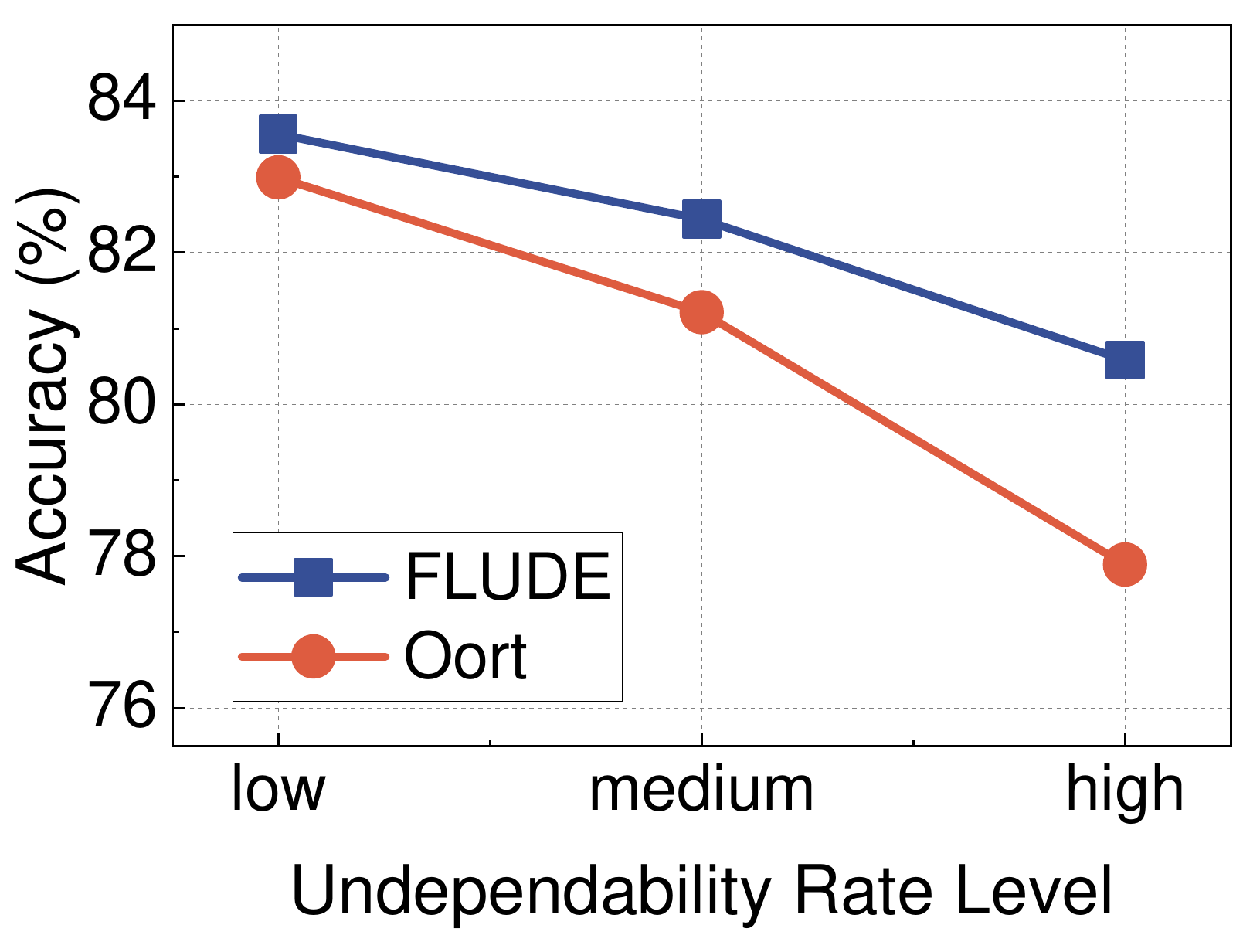}
            }
    \end{minipage}
    \caption{Impact of device undependability rate on final accuracy.}\label{robustness-undependability-accuracy}
\end{figure}

Devices in FL systems will dynamically be online or offline for various reasons. For example, users may choose to kill the FL process on devices if they do not require the services provided by FL. We first examine the impact of the device's online rates on FLUDE performance. 
Specifically, we fix the time interval for state (online or offline) transition as 10 minutes for all devices. Besides, we vary the online rates for devices with \{0.5, 0.3, 0.1\} to simulate the system with low, medium and high levels of offline rates, respectively.
Figure \ref{robustness-dynamics-accuracy-googlespeech} shows that the final accuracy of FLUDE slightly degrades with increasing offline rates. For instance, when the offline rate changes from medium to high, the final accuracy of FLUDE on the Google Speech dataset decreases by only 1.77\%, whereas Oort's accuracy decreases by 3.22\%. These results demonstrate that FLUDE effectively handles scenarios with high offline rates.

Next, we examine the impact of device undependability rates during local training on FLUDE's performance by varying the settings of the device's undependability rate. Specifically, we assume the undependability rates of all devices follow a normal distribution with a variance of 0.05. We vary the mean of the normal distribution with values {0.2, 0.4, 0.6} to simulate systems with low, medium, and high levels of undependability rates, respectively.
As shown in Figure \ref{robustness-undependability-accuracy}, FLUDE consistently achieves higher final accuracy than Oort under different undependability rates on both the CIFAR-100 and Google Speech datasets. Additionally, FLUDE's final accuracy experiences only a slight drop as the undependability rate increases, whereas Oort's final accuracy exhibits a more significant decline. For instance, as depicted in Figure \ref{robustness-undependability-accuracy-cifar100}, FLUDE's final accuracy on CIFAR-100 decreases by 4.57\% when the undependability rate level increases from low to high, while Oort's final accuracy decreases by 8.68\%. These results demonstrate that our proposed framework FLUDE remains robust in highly undependable environments.
\\ 

%% file: content/works.tex
Federated learning (FL) \cite{mcmahan2017communication, beltran2023decentralized, zeng2023fedlab}, epitomized by the FedAvg algorithm \cite{mcmahan2017communication}, represents an emerging distributed machine learning paradigm designed to protect user data privacy. 
FL enables devices distributed across different geographical locations to collaboratively participate in model training while keeping their data local.
While FL inherently offers advantages in safeguarding data privacy, it still faces unique challenges in system efficiency and model performance that are not encountered in traditional distributed machine learning within centralized data centers \cite{khan2021federated, mammen2021federated}.\\
\textbf{Model Performance Optimization in FL.} The problem of optimizing model performance in federated learning can be framed as minimizing an FL loss function that captures the performance of the FL algorithm. Various factors in FL, such as training hyperparameters, the quality of data on participating devices, and data distribution, can significantly impact model performance. 
A vast amount of research \cite{li2020federated, reddiadaptive, zhang2023fedur, zhang2023fedala, wang2024towards, panchal2024flow, yang2024fedas, lai2021oort, li2022pyramidfl} has been dedicated to improving model performance by mitigating the adverse effects of these factors. For example, YoGi \cite{reddiadaptive} and Prox \cite{li2020federated} design specialized federated learning optimizers to overcome the challenges of data heterogeneity. 
Oort \cite{lai2021oort} and PyramidFL \cite{li2022pyramidfl} carefully select devices with high data utility for training to improve final accuracy of the global model. However, these existing works often operate under the assumption of fully dependable environments by default for FL systems, limiting their effectiveness in scenarios characterized by device undependability.\\
\textbf{System Efficiency Optimization in FL.} Communication and computation always represent critical bottlenecks in the efficiency of FL systems. This is primarily due to the frequent transmission of large volumes of model parameters between devices and the central server, coupled with intensive computation demands for model updates on resource-constrained devices \cite{shahid2021communication}. There are quite a large amount of works focusing on reducing communication costs and computational complexity to enhance the efficiency of FL, through techniques such as model quantization/compression \cite{shlezinger2020uveqfed, jhunjhunwala2021adaptive, abdelmoniem2021towards, tonellotto2021neural, amiri2020federated, shah2021model, malekijoo2021fedzip, tang2024bandwidth} and model distillation \cite{li2019fedmd, zhu2021data, wu2022communication, yang2024fedfed, wang2024dfrd}. For example, AdaQuanFL [54] introduces an adaptive model quantization strategy that dynamically adjusts quantization levels throughout the FL cycle, thereby reducing communication overhead while maintaining a low error rate. BCRS [60] dynamically adjusts compression ratios according to available bandwidth on devices, enabling them to upload their models at a close pace, thus effectively exploiting the otherwise wasted time to transmit more data. Furthermore, some contemporary approaches, with Oort \cite{lai2021oort} and PyramidFL \cite{li2022pyramidfl} as typical representatives, strive to simultaneously enhance both system efficiency and model performance within FL systems. Our work shares a similar objective with these works on the joint optimization; however, we tackle this problem from the unique perspective of mitigating the resource wastage and the model performance degradation caused by the presence of numerous undependable devices in FL systems.\\



%% file: content/conclusion.tex
In this paper, we propose an efficient FL framework, named FLUDE, to address the device undependability challenges in FL. 
Specifically, FLUDE introduces several novel techniques to enhance the performance of FL systems. We have implemented FLUDE on two physical platforms with 40 OPPO smartphones and 80 NVIDIA Jetson devices. The experimental results demonstrate the efficiency of FLUDE. 
Considering the unlabeled data in some practical applications, we will further study how to deal with device undependability in semi-supervised federated learning.
